\documentclass{article}
\usepackage{microtype}
\usepackage{graphicx}
\usepackage{subfigure}
\usepackage{booktabs} 
\usepackage{hyperref}

\usepackage[accepted]{icml2023}
\usepackage[T1]{fontenc}
\usepackage{hyperref}
\usepackage{url}
\usepackage{booktabs}       % professional-quality tables
\usepackage{amsfonts}       % blackboard math symbols
\usepackage{nicefrac}       % compact symbols for 1/2, etc.
\usepackage{microtype}      % microtypography
\usepackage{xcolor}         % colors
\usepackage{amsmath}
\usepackage{amssymb}
\usepackage{amsthm}
\usepackage{algorithm}
\usepackage{algorithmic}
\usepackage{xspace}
\usepackage{graphicx}
\usepackage{subfigure}
\usepackage{wrapfig}
\usepackage{graphicx}
\usepackage{subfigure}
\usepackage{url}
\usepackage{color}
\usepackage{multirow}
\usepackage{gensymb}
\usepackage{longtable}
\usepackage{tikz}
\usepackage{comment}
\usepackage{dsfont}
\usepackage{framed}
\usepackage{enumitem}
\usepackage{xspace}
\usepackage{bm}
\usepackage[normalem]{ulem}
\usepackage{booktabs}
\usepackage{vcell}
\usepackage{multibib}
\newcites{citenew}{Reference}
\theoremstyle{plain}
\newtheorem{theorem}{Theorem}[section]
\newtheorem{corollary}{Corollary}[theorem]
\newtheorem{lemma}[theorem]{Lemma}

\theoremstyle{definition}
\newtheorem{definition}{Definition}

\newcommand{\ouralg}{\texttt{LOMAR}\xspace}

\newcommand{\opt}{\text{OPT}\xspace}

\newcommand{\greedy}{\text{Greedy}\xspace}
\newcommand{\ml}{\text{DRL}\xspace}
\newcommand{\rob}{\text{DRL-OS}\xspace}
\newcommand{\osm}{\text{OSM}\xspace}

\begin{document}

\twocolumn[
\icmltitle{Learning for Edge-Weighted Online Bipartite Matching with Robustness Guarantees}

\begin{icmlauthorlist}
\icmlauthor{Pengfei Li}{sch}
\icmlauthor{Jianyi Yang}{sch}
\icmlauthor{Shaolei Ren}{sch}
\end{icmlauthorlist}

\icmlaffiliation{sch}{University of California, Riverside, CA 92521, United States}

\icmlcorrespondingauthor{Shaolei Ren}{sren@ece.ucr.edu}

\icmlkeywords{Machine Learning, ICML}
\vskip 0.3in
]

\printAffiliationsAndNotice{}

\begin{abstract} 
Many problems, such as online ad display, can be formulated as online bipartite matching. The crucial challenge lies in the nature of sequentially-revealed online item information, based on which we make irreversible matching decisions at each step. While numerous expert online algorithms have been proposed with bounded worst-case competitive ratios, they may not offer satisfactory performance in average cases. On the other hand, reinforcement learning (RL) has been applied to improve the average performance, but it lacks robustness and can perform arbitrarily poorly. In this paper, we propose a novel RL-based approach to edge-weighted online bipartite matching with robustness guarantees (\ouralg), achieving both good average-case and worst-case performance. The key novelty of \ouralg is a new online switching operation which, based on a judicious condition to hedge against future uncertainties, decides whether to follow the expert's decision or the RL decision for each online item. We prove that for any $\rho\in[0,1]$, \ouralg is $\rho$-competitive against any given expert online algorithm. To improve the average performance, we train the RL policy by explicitly considering the online switching operation. Finally, we run empirical experiments to demonstrate the advantages of \ouralg compared to existing baselines. Our code is available at:~\href{https://github.com/Ren-Research/LOMAR}{https://github.com/Ren-Research/LOMAR}
\end{abstract}

\section{Introduction}

Online bipartite matching  is
a classic online problem of  practical importance \citep{OnlineMatching_Booklet_OnlineMatchingAdAllocation_Mehta_2013_41870,OBM_EdgeWeighted_NoFreeDisposal_Korea_2020_KIM2020106370,OBM_EdgeWeighted_OBM_ZhiyiHuang_FOCS_2020_9317873,OBM_prediction_antoniadis2020secretary,OBM_StochasticPoisson_ZhiyiHuang_STOC_2021_10.1145/3406325.3451079,OnlineOpt_DataDrivenAlgorithmDesign_Roughgarden_Stanford_CACM_2020_10.1145/3394625}.
In a nutshell, online bipartite matching assigns online items to offline items in two separate sets: when an online item arrives, we need to match it to an offline item given applicable constraints (e.g., capacity constraint),
with the goal of maximizing the total rewards collected  
\citep{OnlineMatching_Booklet_OnlineMatchingAdAllocation_Mehta_2013_41870}.
For example, numerous applications, including scheduling tasks to servers, displaying advertisements to online users, recommending articles/movies/products, among many others, can all be modeled as online bipartite matching or its variants.

The practical importance, along with
substantial algorithmic challenges, of online bipartite matching has
received extensive attention in the last few decades \citep{OBM_Early_FOCS_1990_10.1145/100216.100262,OBM_EdgeWeighted_OBM_ZhiyiHuang_FOCS_2020_9317873}. Concretely,
many algorithms have been proposed and studied for various settings
of online bipartite matching, ranging from simple yet effective greedy algorithms
to sophisticated ranking-based algorithms \citep{OBM_Early_FOCS_1990_10.1145/100216.100262,OBM_EdgeWeighted_NoFreeDisposal_Korea_2020_KIM2020106370,OBM_EdgeWeighted_OBM_ZhiyiHuang_FOCS_2020_9317873,RANKING_aggarwal2011online,ranking_primal_dual_analysis_devanur2013randomized}.
These expert algorithms typically have robustness guarantees in
terms of the competitive ratio --- the ratio of the total reward
obtained by an online algorithm to the reward of another baseline algorithm
(commonly the optimal offline algorithm) --- even under adversarial settings given
arbitrarily bad problem inputs \citep{OBM_Early_FOCS_1990_10.1145/100216.100262,OBM_StochasticPoisson_ZhiyiHuang_STOC_2021_10.1145/3406325.3451079}.  
In some settings, 
even the optimal competitive ratio for adversarial inputs
has been derived (readers are referred to \citep{OnlineMatching_Booklet_OnlineMatchingAdAllocation_Mehta_2013_41870}
for an excellent tutorial).
The abundance of competitive online algorithms has clearly demonstrated the 
importance of performance robustness in terms
of the competitive ratio, especially in safety-sensitive applications such
as matching mission-critical items or under contractual obligations
\citep{OBM_EdgeWeighted_OBM_ZhiyiHuang_FOCS_2020_9317873}. Nonetheless, 
as commonly known in the literature,
the necessity of conservativeness to address the worst-case adversarial input
means that the average performance is typically not optimal 
(see, e.g., \citep{SOCO_ML_ChasingConvexBodiesFunction_Adam_COLT_2022,OnlineOpt_DataCompetitiveKnapsack_Adam_AAAI_2021_DBLP:conf/aaai/Zeynali0HW21} for discussions in other general online problems).

More recently, online optimizers based on reinforcement learning (RL)
\citep{OBM_RL_NonStationary_Distribution_GuihaiChen_ACM_Trans_Knowledge_2022_10.1145/3502734,OBM_ReinforcementLearning_Neural_arXiv_2020_DBLP:journals/corr/abs-2005-11304,OBM_RL_Adaptive_ICDE_NTU_2019_8731455,L2O_OnlineBipartiteMatching_Toronto_ArXiv_2021_DBLP:journals/corr/abs-2109-10380,L2O_OnlineResource_PriceCloud_ChuanWu_AAAI_2019_10.1609/aaai.v33i01.33017570,L2O_Adversarial_Robust_GAN_arXiv_2020}
have been proposed in the context of online bipartite matching as well as other online problems. Specifically, by exploiting statistical information of problem inputs, RL models are trained offline 
and then applied online to produce decisions given unseen problem inputs.
These RL-based optimizers can often achieve high average rewards
in many typical cases. Nonetheless, they may not have
any performance robustness guarantees in terms of the competitive ratio.
In fact, a crucial pain point is that the worst-case performance of many RL-based optimizers
can be arbitrarily bad, due to, e.g., testing distribution shifts,
inevitable model generalization errors, finite samples, and/or even adversarial inputs. Consequently, the lack of robustness guarantees has become a key roadblock for wide deployment of RL-based optimizers
in real-world applications.

In this paper, we focus on an important and novel
objective ---
achieving both good average performance and guaranteed worst-case robustness
--- for \emph{edge-weighted} online bipartite matching \citep{OBM_EdgeWeighted_OBM_ZhiyiHuang_FOCS_2020_9317873,OBM_EdgeWeighted_NoFreeDisposal_Korea_2020_KIM2020106370}.
More specifically, our algorithm, called \ouralg
(Learning-based approach to edge-weighted
Online bipartite MAtching with Robustness guarantees),
 integrates an expert algorithm with  RL.
The key novelty of \ouralg lies in a carefully-designed online \emph{switching} step that dynamically switches between the RL decision and the expert decision online, as well as a switching-aware training algorithm. 
For both no-free-disposal and free-disposal
settings, we design novel
switching conditions as to when the RL decisions can be safely followed while still guaranteeing robustness
of being  $\rho$-competitive  against
\emph{any} given expert online algorithms for any $\rho\in[0,1]$.
To improve the average performance of \ouralg, 
we train the RL policy in \ouralg by explicitly taking
into account the introduced switching operation.
Importantly, to avoid the ``no supervision''
trap during the initial RL policy training, we propose
to approximate the switching operation probabilistically.
Finally, we offer empirical experiments to demonstrate
that \ouralg can improve the average cost (compared
to existing expert algorithms) as well as lower
the competitive ratio (compared to pure RL-based optimizers).

\section{Related Works}\label{sec:related}

Online bipartite matching has
been traditionally approached by expert algorithms \citep{OnlineMatching_Booklet_OnlineMatchingAdAllocation_Mehta_2013_41870,OBM_unkown_distribution_karande2011online,bipatite_matching_huang2019online, ranking_primal_dual_analysis_devanur2013randomized}. A simple but widely-used algorithm is the (deterministic) greedy algorithm \citep{OnlineMatching_Booklet_OnlineMatchingAdAllocation_Mehta_2013_41870}, achieving reasonably-good competitive ratios and empirical performance
\citep{L2O_OnlineBipartiteMatching_Toronto_ArXiv_2021_DBLP:journals/corr/abs-2109-10380}.
Randomized algorithms have also been proposed to improve the competitive ratio \citep{greedy_rt_ting2014near,RANKING_aggarwal2011online}. 
 In addition,  competitive algorithms based on the primal-dual framework have also been proposed \citep{OnlineMatching_Booklet_OnlineMatchingAdAllocation_Mehta_2013_41870,primal_dual_online_buchbinder2009design}.
More recently, 
  multi-phase information and predictions have
  been leveraged to exploit stochasticity within each problem instance
  and improve the algorithm performance \citep{kesselheim2013optimal}.
For example,  
\citep{secretary_matching_korula2009algorithms} designs a secretary matching algorithm based on a threshold obtained using the information of phase one, and exploits the threshold for matching in phase two.
Note that stochastic settings considered by expert algorithms \citep{OnlineMatching_Booklet_OnlineMatchingAdAllocation_Mehta_2013_41870,OBM_unkown_distribution_karande2011online} mean that
the arrival orders and/or rewards of different online items within each problem instance are stochastic.
By contrast, as shown in \eqref{objective}, we focus on an unknown
distribution of problem instances whereas the inputs within each instance can still be arbitrary.

Another line of algorithms utilize RL to improve the average performance \citep{OBM_RL_Adaptive_ICDE_NTU_2019_8731455,OBM_ReinforcementLearning_Neural_arXiv_2020_DBLP:journals/corr/abs-2005-11304,OBM_RL_NonStationary_Distribution_GuihaiChen_ACM_Trans_Knowledge_2022_10.1145/3502734, L2O_OnlineBipartiteMatching_Toronto_ArXiv_2021_DBLP:journals/corr/abs-2109-10380}.
Even though heuristic methods (such as using adversarial 
training samples \citep{L2O_Adversarial_Robust_GAN_arXiv_2020,L2O_AdversarialOnlineResource_ChuanWu_HKU_TOMPECS_2021_10.1145/3494526})
are used to empirically improve the robustness,
they do not provide any theoretically-proved robustness guarantees.

ML-augmented algorithms have been recently considered for various problems
\citep{SOCO_OnlineOpt_UnreliablePrediction_Adam_Sigmetrics_2023_10.1145/3579442,OBM_RobustnessConsistency_TwoStage_NeurIPS22_jin2022online,SOCO_ML_ChasingConvexBodiesFunction_Adam_COLT_2022,OnlineOpt_ML_Augmented_RobustCache_Google_ICML_2021_pmlr-v139-chledowski21a,OnlineOpt_ML_Advice_CompetitiveCache_Google_JACM_2021_10.1145/3447579}. 
By viewing the ML prediction as 
blackbox advice, these algorithms strive to provide good competitive ratios when the ML predictions are nearly perfect,
and also bounded competitive ratios when ML predictions are bad. 
 But, they still focus on the worst case without addressing
 the average performance or how the ML model is trained.
By contrast, the RL model in \ouralg is trained by taking into account the switching operation and performs inference
based on the actual state (rather than 
its own independently-maintained state as a blackbox).
 Assuming a given downstream algorithm,
\citep{L2O_PredictOptimize_MDP_Harvard_NIPS_2021_wang2021learning,L2O_PredictOptimize_RiskCalibrationBound_ICLR_2021_liu2021risk,decision_focused_learning_combinatorial_opt_wilder2019melding,L2O_PredictOptimize_arXiv_2017,L2O_LearningMLAugmentedAlgorithm_Harvard_ICML_2021_pmlr-v139-du21d,L2O_LearningMLAugmented_Regression_CR_GeRong_NIPS_2021_anand2021a} 
focus on learning the ML model 
to better serve the end goal in completely different (sometimes, offline optimization) problems.

\ouralg  is relevant
to conservative  bandits/RL \citep{Conservative_Bandits_ICML_2016_10.5555/3045390.3045523,Conservative_Bandits_LinearContextual_NIPS2017_bdc4626a,Conservative_RL_Bandits_LiweiWang_SimonDu_ICLR_2022_yang2022a,Conservative_RL_Facebook_AISTATS_2020_pmlr-v108-garcelon20a}. With unknown
reward functions (as well as transition models if applicable),
conservative  bandits/RL leverages
an existing policy to safeguard the exploration process. But, 
they only consider the cumulative reward without addressing future uncertainties
when deciding exploration vs. rolling
back to an existing policy.
 Thus, as shown in Section~\ref{sec:algorithm},
this cannot guarantee robustness 
 in our problem.
Also, constrained policy optimization
 \citep{Conservative_ProjectBasedConstrainedPolicyOptimizatino_ICLR_2020_Yang2020Projection-Based,Conservative_QLearning_OffilineRL_Berkeley_NIPS_2020_NEURIPS2020_0d2b2061,Conservative_TrustRegionPolicyOptimization_ICML_2015_pmlr-v37-schulman15,Conservative_ConstrainedPolicyOptimization_achiam2017constrained,Safe_RL_ImagineFutureStates_TengyuMa_NIPS_2021_NEURIPS2021_73b277c1,Safe_RL_ModelBased_Stability_Krause_ETHZ_NIPS_2017_10.5555/3294771.3294858}
focuses on  average (cost) constraints in the long run, 
whereas  \ouralg achieves stronger robustness (relative to an expert algorithm) for any episode.

\section{Problem Formulation}

We focus on  \emph{edge-weighted} online bipartite matching, which includes un-weighted and vertex-weighted matching as special cases \citep{OBM_EdgeWeighted_OBM_ZhiyiHuang_FOCS_2020_9317873,OBM_EdgeWeighted_NoFreeDisposal_Korea_2020_KIM2020106370}. 
In the following, we
also drop ``edge-weighted'' if applicable when referring to our problem. 

\subsection{Model}
The  agent matches  items (a.k.a. vertices) between two sets $\mathcal{U}$ and $\mathcal{V}$ to gain as high total rewards as possible. Suppose that $\mathcal{U}$ is fixed 
and contains \emph{offline} items
$u\in\mathcal{U}$,
and that the \emph{online} items  $v\in\mathcal{V}$ arrive sequentially: in each time slot, an online item $v\in\mathcal{V}$ arrives and the weight/reward information $\{w_{uv}\mid w_{u,\min} \leq w_{uv}\leq w_{u,\max}, u\in \mathcal{U}\}$ is revealed,
where
$w_{uv}$ represents the reward 
when the online item $v$ is matched to each offline $u\in \mathcal{U}$. 
We denote one problem instance by $\mathcal{G}=\left\{\mathcal{U},\mathcal{V},\mathcal{W}\right\}$, where
$\mathcal{W}=\{w_{uv}\mid u\in \mathcal{U}, v\in \mathcal{V}\}$.
%For online decisions, the agent does \emph{not} know the reward of future online items.
%Also, any offline item $u\in\mathcal{U}$ can be matched up to $c_u$ times, where $c_u$ is essentially the capacity for offline item $u$ known to the agent. 
 We denote $x_{uv}\in\{0,1\}$ as the matching decision indicating whether 
$u$ is matched to $v$.  Also, any offline item $u\in\mathcal{U}$ can be matched up to $c_u$ times, where $c_u$ is essentially the capacity for offline item $u$ known to the agent in advance.

The goal is to maximize the total collected reward $\sum_{v\in\mathcal{V},u\in\mathcal{U}}x_{uv}w_{uv}$. With a slight abuse of notations, we denote $x_v\in\mathcal{U}$ as the index of item in $\mathcal{U}$ that is matched to item $v\in\mathcal{V}$. The set of
online items matched to $u\in\mathcal{U}$ is denoted as
$\mathcal{V}_u=\{v\in\mathcal{V}\,|\, x_{uv}=1\}$.

The edge-weighted online bipartite matching problem
has been mostly studied under two different settings:
no free disposal and with free disposal \citep{OnlineMatching_Booklet_OnlineMatchingAdAllocation_Mehta_2013_41870}. In the no-free-disposal
case, each offline item $u\in\mathcal{U}$ can only be matched
strictly up to $c_u$ times;
in the free-disposal case, 
each offline item $u\in\mathcal{U}$ can be matched more than $c_u$ times, but
only the top $c_u$ rewards are counted
when more than $c_u$ online items
are matched to $u$. 
Compared to the free-disposal case,
the no-free-disposal case is significantly more challenging with the optimal competitive ratio being
$0$ in the strong adversarial setting unless 
additional assumptions are made (e.g., $w_{u,\min}>0$ for each $u\in\mathcal{U}$ \citep{OBM_EdgeWeighted_NoFreeDisposal_Korea_2020_KIM2020106370} and/or random-order of online arrivals) \citep{OBM_EdgeWeighted_OBM_ZhiyiHuang_FOCS_2020_9317873,OnlineMatching_Booklet_OnlineMatchingAdAllocation_Mehta_2013_41870}. 
The free-disposal
setting is not only analytically more tractable,
but also is practically motivated by the display ad application where
the advertisers (i.e., offline items $u\in\mathcal{U}$)
will not be unhappy if they receive more impressions (i.e., online
items $v\in\mathcal{V}$) than their budgets $c_u$, even though only
the top $c_u$ items count.

 \ouralg can handle both
no-free-disposal and free-disposal settings. 
For better presentation of our key novelty
and page limits,
 we focus on the no-free-disposal setting in the body
 of the paper, \emph{while deferring the free-disposal setting
 to Appendix~\ref{appendix_free_disposal}}.

Specifically, 
the \textbf{offline} problem with no free disposal can be expressed as:
\begin{equation}\label{eqn:offline_problem}
	\begin{gathered}   
\max\sum_{x_{uv}\in\{0,1\}, u\in\mathcal{U},v\in\mathcal{V}}x_{uv}w_{uv},\\
	\;\;\mathrm{s.t.},\;\;   \sum_{v\in\mathcal{V}}x_{uv}\leq c_u,
 \text{ and }	 \sum_{u\in\mathcal{U}}x_{uv}\leq 1, \forall u\in\mathcal{U}, v\in\mathcal{V}
	\end{gathered}
\end{equation}
where the constraints specify
the offline item capacity limit and 
each online item $v\in{\mathcal{V}}$
can only be matched up to one offline item $u\in\mathcal{U}$.
Given an online algorithm $\alpha$, 
we use
$f^{\alpha}_u(\mathcal{G})$ to denote the total reward collected for offline item $u\in\mathcal{U}$, and $R^{\alpha}(\mathcal{G})=\sum_{u\in\mathcal{U}}f^{\alpha}_u(\mathcal{G})$ to denote
the total collected reward. We will also drop the superscript
$\alpha$ for notational convenience wherever applicable.

\subsection{Objective}
Solving the problem in \eqref{eqn:offline_problem} is very challenging
in the online case, where
 %In the online case, 
the agent has
to make irreversible decisions
without knowing the future online item
arrivals. 
Next, we define a generalized competitiveness
as a metric of robustness and then present our optimization objective.

\begin{definition}[Competitiveness]
An online bipartite matching algorithm $\alpha$
is said to be $\rho$-competitive with  $\rho\
\geq 0$
against the algorithm $\pi$ 
if for any problem instance $\mathcal{G}$,
its total collected reward $R^{\alpha}(\mathcal{G})$ satisfies 
$R^{\alpha}(\mathcal{G})\geq \rho R^{\pi}(\mathcal{G})-B$,
where $B\geq 0$ is a constant independent of the problem input, and $R^{\pi}$ is the total reward  of the algorithm $\pi$. 
\end{definition}

Competitiveness against a given online algorithm $\pi$ (a.k.a., expert) is common in
the literature on algorithm designs \citep{SOCO_ML_ChasingConvexBodiesFunction_Adam_COLT_2022}:
the greater $\rho\geq0$, the better robustness of the online algorithm, although the average rewards can be worse.
The constant $B\geq0$
relaxes the strict competitive ratio by allowing an additive \emph{regret} \citep{SOCO_MetricUntrustedPrediction_Google_ICML_2020_pmlr-v119-antoniadis20a}.
When $B=0$, the competitive ratio  becomes the strict one.
In practice, the expert algorithm $\pi$ can be viewed
as an existing solution currently in use, while the new 
RL-based algorithm is being pursued subject to a constraint
that the collected reward must be at least $\rho$ times of the expert.
Additionally, if the expert itself has
a competitive ratio of $\lambda\leq1$ against the offline oracle algorithm (\opt), then
it will naturally translate into \ouralg being
$\rho\lambda$-competitive against \opt.

On top of worst-case robustness, we are also interested
in the average award.
Specifically, we focus on a
setting where the problem instance
$\mathcal{G}=\left\{\mathcal{U},\mathcal{V},\mathcal{W}\right\}$
follows an \emph{unknown} distribution, whereas
both the rewards $\mathcal{W}$ and online arrival order
within each instance $\mathcal{G}$ can be adversarial. 

Nonetheless, the average reward and worst-case robustness are different,
and optimizing one metric alone does not necessarily optimize the other one (which
is a direct byproduct of  Yao's principle \cite{YaoPrinciple_UnifiedMeasureComplexity_FOCS_1977_10.1109/SFCS.1977.24}.
In fact, there is a tradeoff between
the average performance and worst-case robustness
in general online problems
 \cite{SOCO_ML_ChasingConvexBodiesFunction_Adam_COLT_2022}.
 The reason is that an
online algorithm that maximizes the average reward 
prioritizes typical problem instances, while
conservativeness is needed by 
a robust algorithm to mitigate the worst-case uncertainties and outliers.

In \ouralg, we aim to maximize  the average reward subject to worst-case robustness
guarantees as formalized below: 
\begin{subequations}\label{objective}
\begin{align}\label{eqn:objective_expected}
&\max \mathbb{E}_{\mathcal{G}}\left[R^{\alpha}(\mathcal{G})\right]\\
\label{eqn:constraint}
\mathrm{s.t.}\;\; & R^{\alpha}(\mathcal{G})\geq \rho R^{\pi}(\mathcal{G})-B,\;\;\;\forall \mathcal{G},
\end{align}
\end{subequations}
where the expectation $\mathbb{E}_{\mathcal{G}}\left[R^{\alpha}(\mathcal{G})\right]$
is over the randomness $\mathcal{G}=\left\{\mathcal{U},\mathcal{V},\mathcal{W}\right\}$.
The worst-case robustness
constraint for each problem instance is significantly more challenging than an average reward constraint.
Our problem in \eqref{objective} is novel in
that it generalizes
the recent RL-based online algorithms \cite{L2O_OnlineBipartiteMatching_Toronto_ArXiv_2021_DBLP:journals/corr/abs-2109-10380} by guaranteeing worst-case robustness;
it leverages robustness-aware RL training (Section~\ref{sec:training}) 
to improve the average reward
and hence
also differs from the prior ML-augmented algorithms
that still predominantly focus on the worst-case performance \cite{SOCO_ML_ChasingConvexBodiesFunction_Adam_COLT_2022,OnlineOpt_Learning_Augmented_RobustnessConsistency_NIPS_2020}.

Some manually-designed algorithms focus
on a \emph{stochastic} setting where the arrival order is random
and/or the rewards $\{w_{uv}\mid w_{u,\min} \leq w_{uv}\leq w_{u,\max}, u\in \mathcal{U}\}$ of each online item is independently and identically distributed
(i.i.d.) within each problem instance $\mathcal{G}$ \cite{OnlineMatching_Booklet_OnlineMatchingAdAllocation_Mehta_2013_41870}.
By contrast, our settings are significantly different ---
we only assume an unknown distribution for the entire
problem instance $\mathcal{G}=\left\{\mathcal{U},\mathcal{V},\mathcal{W}\right\}$
while  both the rewards $\mathcal{W}$ and online arrival order
within each instance $\mathcal{G}$ can be arbitrary.

\section{Design of Online Switching for Robustness% Guarantees
}\label{sec:algorithm}

Assuming that the RL policy is already trained (to be addressed
in Section~\ref{sec:training}),
we now present the inference of \ouralg, which includes
novel online \emph{switching} to
dynamically follow the RL decision or the expert decision,
for robustness guarantees against the
expert.

\subsection{Online Switching}

\begin{algorithm}[!t]
	\caption{Inference of
	 Robust Learning-based  Online Bipartite Matching (\ouralg)} 
	\begin{algorithmic}[1]\label{alg:1}
 \REQUIRE Competitiveness constraint $\rho\in[0,1]$ and $B\geq0$
		\FOR {$v= 1$ to $|\mathcal{V}|$}
		 \STATE   Run the expert $\pi$ and get expert's decision $x_{v}^{\pi}$. 
		\STATE   
  If $x_{v}^{\pi}\neq \mathrm{skip}$:
$\mathcal{V}_{x_{v}^{\pi},v}^{\pi}=
		\mathcal{V}_{x_{v}^{\pi},v-1}^{\pi} \bigcup\{v\}$, \\
  $R^{\pi}_v=R^{\pi}_{v-1}+w_{x_v^{\pi},v}$.\\
  \texttt{//Update the virtual decision set and reward of  the expert}
		\STATE ${s}_u=w_{uv}-h_{\theta}(I_{u},	w_{uv})$, $\forall u\in\mathcal{U}$\\ 
  \texttt{//Run RL model to get score $s_u$ with history information $I_{u}$}
  \STATE 
  $\tilde{x}_v=\arg \max_{u\in\mathcal{U}_a\bigcup \{\mathrm{skip}\}} \bigl\lbrace \{s_u\}_{u\in\mathcal{U}_a}, s_{\mathrm{skip}} \bigr\rbrace$, with $s_{\mathrm{skip}}=0$ and $\mathcal{U}_a=\left\lbrace u\in \mathcal{U}\mid |\mathcal{V}_{u,v-1}|< c_u\right\rbrace$.\\
    \texttt{//Get RL decision $\tilde{x}_v$}
       \IF {Robust constraint in \eqref{eqn:condition} is satisfied}\label{step:switchbegin}
		\STATE Select $x_v=\tilde{x}_v$. \texttt{//Follow RL}
		\ELSIF {$x^{\pi}_{v}$ is available  (i.e., $|\mathcal{V}_{x^{\pi}_v,v-1}|<c_{x^{\pi}_v}$)}
		\STATE Select $x_v=x^{\pi}_{v}$.    \texttt{//Follow the expert}
		\ELSE
		\STATE Select $x_v=\mathrm{skip}$.
		\ENDIF\label{step:switchend}
		\STATE  If $x_v\neq\mathrm{skip}$, $\mathcal{V}_{x_{v},v}=
		\mathcal{V}_{x_{v},v-1}\bigcup\{v\}$,\\ $R_v=R_{v-1}+w_{x_v,v}$.\\
       \texttt{//Update the true decision set and reward}
		\ENDFOR
	\end{algorithmic}
\end{algorithm}

While switching is common in (ML-augmented) online algorithms, ``\emph{how to switch}'' is highly non-trivial and a key merit for algorithm designs \cite{SOCO_MetricUntrustedPrediction_Google_ICML_2020_pmlr-v119-antoniadis20a,SOCO_ML_ChasingConvexBodiesFunction_Adam_COLT_2022,SOCO_OnlineOpt_UnreliablePrediction_Adam_Sigmetrics_2023_10.1145/3579442}. 
 To  
guarantee robustness (i.e., $\rho$-competitive
against a given expert for any $\rho\in[0,1]$),
we propose a novel online  algorithm (Algorithm~\ref{alg:1}).
In the algorithm, we independently run an expert online algorithm 
$\pi$  --- the cumulative reward and
item matching decisions are all maintained virtually for the expert,
but not used as the actual decisions. Based on the performance of expert online algorithm, we design a robust constraint which serves as the condition for online switching.

Concretely, we define the set of items that is actually matched to offline item $u\in\mathcal{U}$ before the start of ${(v+1)}-$th step as $\mathcal{V}_{u,v}$, and the set of items that is virtually matched to offline item $u\in\mathcal{U}$ by expert before the start of ${(v+1)}-$th step as $\mathcal{V}^{\pi}_{u,v}$. Initially, we have $\mathcal{V}_{u,0}=\emptyset$, and $\mathcal{V}^{\pi}_{u,0}=\emptyset$. We also denote $\mathcal{U}_a$ as the set of available offline item and initialize it as $\mathcal{U}$. When an online item $v$ arrives at each step, Algorithm~\ref{alg:1} first runs the expert algorithm $\pi$, gets the expert decision $x_v^{\pi}$ and update the virtual decision set and reward if the expert decision is not skipping this step. Then the RL policy gives the scores $s_u$ of each offline item $u\in\mathcal{U}$. By assigning the score of skipping as $0$ and comparing the scores, the algorithm obtain the RL action advice $\tilde{x}_v$. Then the algorithm perform online switching to guarantee robustness.

The most crucial step for safeguarding
RL decisions is our  online {switching} step: Lines~\ref{step:switchbegin}--\ref{step:switchend} in Algorithm~\ref{alg:1}. 
The key idea for this step
is to switch
between the expert decision $x_v^{\pi}$ and the RL decision
$\tilde{x}_v$ in order to ensure that the actual online decision $x_{v}$
meets the $\rho$-competitive requirement (against the expert $\pi$).
Specifically, we follow the RL decision $\tilde{x}_v$ only
if it can safely hedge against any future uncertainties
(i.e., the expert's future reward increase); otherwise, we
need to roll back to the expert's decision $x_v^{\pi}$ to stay on
track for robustness.

Nonetheless, naive switching conditions, e.g.,
only ensuring that the actual cumulative reward is at least
$\rho$ times of the expert's cumulative reward at each step \citep{Conservative_Bandits_ICML_2016_10.5555/3045390.3045523,Conservative_RL_Bandits_LiweiWang_SimonDu_ICLR_2022_yang2022a},
can fail to meet the competitive ratio requirement in the end. The reason is that,
even though the competitive ratio requirement is met (i.e.,
$R_v\geq \rho R_v^{\pi}-B$) at the current step $v$, the expert can possibly obtain much higher rewards from future online items $v+1,v+2,\cdots$, 
if it has additional offline item capacity that the actual algorithm \ouralg does not have. 
Thus, we must carefully design the switching conditions to hedge against future risks.

\subsection{Robustness Constraint}

In the no-free-disposal case, an offline item $u\in\mathcal{U}$ cannot receive any additional online items if it has been matched for $c_u$ times up to its capacity. By assigning more online items to  $u\in\mathcal{U}$ than the expert algorithm at step $v$, \ouralg can possibly receive a higher cumulative reward 
than the expert's cumulative reward. But,
such advantages are just \emph{temporary}, because
the expert may receive an even higher reward in the future by filling up the unused capacity of item $u$.
Thus, to hedge against the future uncertainties, 
\ouralg chooses the RL decisions only when the following condition is satified:
\begin{equation}\label{eqn:condition}
\begin{aligned}
R_{v-1}+w_{\tilde{x}_v,v} \geq& \rho \Bigl( R^{\pi}_v \sum_{u\in \mathcal{U}}\bigl(|\mathcal{V}_{u,v-1}|-|V_{u,v}^{\pi}| \\
&+\mathbb{I}_{u=\tilde{x}_v}\bigr)^+\cdot w_{u,\max}\Bigl)-B,
\end{aligned}
\end{equation}
where $\mathbb{I}_{u=\tilde{x}_v}=1$
if and only if $u=\tilde{x}_v$ and 0 otherwise, $(\cdot)^+=\max(\cdot,0)$,
$\rho\in[0,1]$ and $B\geq 0$ are the hyperparameters indicating the desired robustness with respect to the expert algorithm $\pi$. 

The interpretation of ~\eqref{eqn:condition} is as follows. The left-hand side is
the total reward of \ouralg after assigning the online item $v$ based on the RL decision (i.e. $\tilde{x}_t$). 
The right-hand side is the expert's cumulative cost $R_v^\pi$, plus the term
$\sum_{u\in \mathcal{U}}\left(|\mathcal{V}_{u,v-1}|-|V_{u,v}^{\pi}|+\mathbb{I}_{u=\tilde{x}_v}\right)^+\cdot w_{u,\max}$ which 
indicates the maximum reward that can be possibly received by the expert in the future. This reservation term is crucial, especially when the expert has more unused capacity than \ouralg. Specifically, 
$|\mathcal{V}_{u,v-1}|$ is the number of online items (after assigning $v-1$ items) already assigned to the offline item $u\in\mathcal{U}$, and hence 
$\left(|\mathcal{V}_{u,v-1}|-|V_{u,v}^{\pi}|+\mathbb{I}_{u=\tilde{x}_v}\right)^+$ represents the number of more online items that \ouralg has assigned to $u$ than the expert if \ouralg follows the RL decision at step $v$. If
\ouralg assigns fewer items than the expert for an offline item
$u\in\mathcal{U}$, there is no need for any hedging because  \ouralg is guaranteed to receive more rewards by filling up the item $u$ up to the expert's assignment level. 

The term $w_{u,\max}$ in ~\eqref{eqn:condition}
is the set as the maximum possible reward for each decision.
Even when $w_{u,\max}$ is unknown in advance, \ouralg still applies
by simply setting $w_{u,\max}=\infty$. In this case,  
\ouralg will be less ``greedy'' than the expert
and never use more resources than the expert at any step. 

While we have focused on the no-free-disposal setting to
highlight the key idea of our switching condition (i.e.,
not following the RL decisions too aggressively by
hedging against future reward uncertainties),
the free-disposal setting requires a very different switching condition,
which we defer to Appendix~\ref{appendix_free_disposal} due to the page limit.

\subsection{Robustness Analysis}\label{sec:analysis}

We now formally show the competitive ratio of \ouralg. The proof is available in the appendix.

\begin{theorem}\label{thm:cr}
For any $0\leq\rho\leq1$ and $B\geq0$ and any expert algorithm $\pi$, 
\ouralg achieves a competitive ratio of $\rho$ against the algorithm $\pi$, i.e.,
$R\geq \rho R^{\pi}-B$ for any problem input.
\end{theorem}

The hyperparameters $0\leq\rho\leq1$ and $B\geq0$ govern
the level of robustness we would like to achieve, at the potential expense
of average reward performance. For example, by setting $\rho=1$ and $B=0$,
we achieve the same robustness as the expert
but leave  little to no freedom for RL decisions.
On the other hand, by setting a small $\rho>0$ and/or large $B$, we provide higher flexibility
to RL decisions for better average performance, while potentially decreasing
the robustness.
In fact, such tradeoff is necessary in the broad context of ML-augmented online algorithms
\citep{SOCO_OnlineOpt_UntrustedPredictions_Switching_Adam_arXiv_2022,SOCO_ML_ChasingConvexBodiesFunction_Adam_COLT_2022}. 
Additionally, in case of multiple experts, we can first combine
these experts into a single expert and then apply \ouralg as
if it works with a single combined expert.

While the competitive ratio of
all online algorithms against the optimal offline algorithm is zero in the no-free-disposal 
and general adversarial setting, there exist provably competitive online expert algorithms
under some technical assumptions and other settings \citep{OnlineMatching_Booklet_OnlineMatchingAdAllocation_Mehta_2013_41870}. For example, 
the simple greedy algorithm
achieves $\left({1+\max_{u\in\mathcal{U}}\frac{w_{u,\max}}{w_{u,\min}}}\right)^{-1}$ under bounded weights assumptions for the adversarial no-free-disposal setting \citep{OBM_EdgeWeighted_NoFreeDisposal_Korea_2020_KIM2020106370}, and $\frac{1}{2}$ for the free-disposal setting \citep{OBM_EdgeWeighted_OBM_ZhiyiHuang_FOCS_2020_9317873}, and there also exist $1/e$-competitive algorithms
against the optimal offline algorithm for the
random-order setting \citep{OnlineMatching_Booklet_OnlineMatchingAdAllocation_Mehta_2013_41870}.
Thus, an immediate result follows.
\begin{corollary}\label{thm:cr_opt}
For any $0\leq\rho\leq1$ and $B\geq0$,
by using Algorithm~\ref{alg:1} and an expert online algorithm $\pi$
that is $\lambda$-competitive against the optimal offline algorithm OPT, 
then under the same assumptions for $\pi$ to be $\lambda $-competitive,
\ouralg is $\rho\lambda$-competitive against OPT.
\end{corollary}

Corollary~\ref{thm:cr_opt} provides
a general result that applies to any
$\lambda$-competitive expert algorithm
$\pi$ under its respective required assumptions. For example,
if the expert $\pi$ assumes an adversarial or random-order setting,
then Corollary~\ref{thm:cr_opt} holds under the same adversarial or random-order setting.

\section{RL Policy Training with Online Switching}\label{sec:training}

The prior ML-augmented online algorithms typically
assume a standalone RL model that is pre-trained
without considering what the online algorithm will perform
\citep{SOCO_ML_ChasingConvexBodiesFunction_Adam_COLT_2022}.
Thus, while the standalone RL model may perform well on its own, 
its performance can be poor when directly used in \ouralg
due to the added online switching step.
In other words, 
there will be a training-testing mismatch.
To rectify the mismatch, we propose a novel approach to train the RL model in \ouralg by explicitly considering the switching operation. 

\textbf{RL architecture.} 
For  online bipartite matching,
there exist various network architectures, e.g., fully-connected networks and scalable invariant network
for general graph sizes. 
The recent study \citep{L2O_OnlineBipartiteMatching_Toronto_ArXiv_2021_DBLP:journals/corr/abs-2109-10380}
has shown using extensive empirical experiments that the invariant  network architecture, where each offline-online item pair runs a separate neural network with shared weights
among all the item pairs, is empirically advantageous, due to its scalability to large graphs and good average performance. 

We denote the RL model as $h_{\theta}(I_u,w_{uv})$
where $\theta$ is the network parameter.
By feeding the item weight $w_{uv}$ and
 applicable history information $I_u$ for
 each offline-online item pair $(u,v)$,
 we can use the RL model to output a \emph{threshold} for possible item assignment,
 following threshold-based algorithms \citep{L2O_OnlineBipartiteMatching_Toronto_ArXiv_2021_DBLP:journals/corr/abs-2109-10380,bipatite_matching_huang2019online,OnlineMatching_Booklet_OnlineMatchingAdAllocation_Mehta_2013_41870}.
 The history information $I_u$
includes, but is not limited to, the average value and variance of weights assigned to $u$, average in-degree of $u$, and maximum weight for the already matched items. More details about the information can be found in the appendix. 
Then, with the RL output, we obtain
a score
$s_u=w_{uv}-h_{\theta}(I_u,w_{uv})$
for each possible assignment.

\textbf{Policy training.} Training
the RL model by considering switching in Algorithm~\ref{alg:1} is  non-trivial.
Most critically, the initial RL decisions can perform arbitrarily badly upon policy initialization,
which means that the initial RL decisions are almost always
overridden by the expert's decisions for robustness.
Due to following the expert's decisions, the RL agent almost always receive a good reward, which actually has nothing to do with the RL's own decisions and hence provides little to no supervision to improve the RL policy. 
Consequently, this 
creates a \emph{gridlock} for RL policy training. While using
an offline pre-trained standalone RL model without considering online switching
(e.g., \citep{L2O_OnlineBipartiteMatching_Toronto_ArXiv_2021_DBLP:journals/corr/abs-2109-10380})
as an initial policy may
partially address this gridlock, this is certainly inefficient as we
have to spend resources for training another RL model, let alone the likelihood
of being trapped into the standalone RL model's suboptimal policies (e.g. local minimums).

\begin{algorithm}[t]
\caption{Policy Training with Online Switching}
\begin{algorithmic}[1]\label{alg:training_brief}
\REQUIRE Competitiveness constraint $\rho\in[0,1]$ and $B\geq0$, initial model weight $\theta$ of RL model.
\FOR {$i= 1$ to n}
\FOR {$v= 1$ to $|\mathcal{V}|$}
\STATE Calculate the actual item selection probability $p_{\theta}(x_{v}|I_{u})$ from Eqn.~\eqref{eqn:actual_probability}. 
\STATE Sample from $p_{\theta}(x_{v}|I_{u})$ to get the item selection $x_v$, then collect the reward $w_{x_v,v}$ for item $v$.
\STATE Update the capacity of the offline item $x_v$ after the assignment $\mathcal{V}_{x_{v},v}$. 
\ENDFOR
\STATE Collect node matching results for the problem instance and add them into a trajectory set.
\ENDFOR
\STATE Estimate policy gradient $\nabla_{\theta}\hat{R}_{\theta}$ based on Eqn.~\eqref{eqn:rl_gradient}.
\STATE Update the RL model weight $\theta$ with $\theta = \theta+\alpha\nabla_{\theta}\hat{R}_{\theta}$;
\end{algorithmic}
\end{algorithm}

\begin{table*}[htp]
\centering
\small
\begin{tabular}{l|ll|ll|ll|ll|ll} 
\toprule
\multicolumn{1}{l}{}  & \multicolumn{2}{c}{\rob}                    & \multicolumn{2}{c}{\ouralg ($\rho=0.4$)}                & \multicolumn{2}{c}{\ouralg ($\rho=0.6$)}                & \multicolumn{2}{l}{\ouralg ($\rho=0.8$)} & \multicolumn{2}{c}{\greedy}                        \\ 
\hline
\multicolumn{1}{c|}{Test} & \multicolumn{1}{c}{AVG} & \multicolumn{1}{c|}{CR} & \multicolumn{1}{c}{AVG} & \multicolumn{1}{c|}{CR} & \multicolumn{1}{c}{AVG} & \multicolumn{1}{c|}{CR} & AVG    & CR                        & \multicolumn{1}{c}{AVG} & \multicolumn{1}{c}{CR}  \\ 
\hline
$\rho = 0.4$          & 12.315       & 0.800        & \textbf{12.364}       & \textbf{0.819}        & 12.288       & 0.804        & 12.284 & 0.804          & 11.000       & 0.723        \\
$\rho = 0.6$& 11.919       & 0.787        & 11.982       & 0.807        & \textbf{11.990}       & \textbf{0.807}        & 11.989 & 0.800          & 11.000       & 0.723        \\
$\rho = 0.8$& 11.524       & \textbf{0.773}       & 11.538       & 0.766        & 11.543       & 0.762        & \textbf{11.561} & 0.765          & 11.000       & 0.723        \\
\bottomrule
\end{tabular}
\vspace{-0.3cm}
\caption{Comparison under different  $\rho$. 
In the top, \ouralg ($\rho=x$) means \ouralg is trained with the value of $\rho=x$. 
The average reward and competitive ratio are represented by AVG and CR, respectively --- the higher, the better. The highest value in each testing setup is highlighted in bold. The AVG and CR for \ml are \textbf{12.909} and \textbf{0.544} respectively. The average reward for \opt is \textbf{13.209}. 
}
\label{table:rho}
\end{table*}

To address these issues, during training, we introduce a softmax probability
 to approximate the otherwise non-differentiable switching operation.
Specifically, the switching probability depends on the cumulative reward difference $R_{diff}$ in the switching condition, which is
\begin{equation}\nonumber
\label{eqn:prob_update}
\begin{aligned}
    R_{diff} = & R_{v-1}+w_{\tilde{x}_v,v}+ B   - \rho \cdot\Bigl( R^{\pi}_v +\\
    &\sum_{u\in \mathcal{U}}\left(|\mathcal{V}_{u,v-1}|-|V_{u,v}^{\pi}|+\mathbb{I}_{u=\tilde{x}_v}\right)^+\cdot w_{u,\max} \Bigr) 
\end{aligned}
\end{equation}
Then, the probability of following RL is $p_{os} = \frac{e^{R_{diff}/t}}{1+ e^{R_{diff}/t}}$, where $t$ is the softmax function's temperature. 
Importantly, this softmax probability is differentiable
and hence allows backpropagation
to train the RL model weight $\theta$ to maximize the expected total reward
while being aware of the switching operation for robustness. 
Next, with \emph{differentiable} switching, we train the RL model by policy gradient \citep{williams1992simple} to optimize the policy parameter $\theta$. Denote $\tau=\{x_{1},\cdots, x_{v}\}$ as an action trajectory sample and $p_{\theta}(x_{v}| I_{u})$ as the probability of matching offline item $u$ to online item $v$: \begin{equation}\label{eqn:actual_probability}
    p_{\theta}(x_{v}|I_{u}) = (1 - p_{os})\cdot \tilde{p}_{\theta}(x_{v}| I_{u}) + p_{os} \cdot p_{\theta}^{\pi}(x_{v}| I_{u}),
\end{equation}
where $\tilde{p}_{\theta}(x_{v}| I_{u})$ is the RL's item selection probability obtained with the RL's output score $s_u$, and $p_{\theta}^{\pi}(x_{v}| I_{u})$ is the item selection probability for expert $\pi$. If the expert's item selection
is not available (i.e., Line~11 in Algorithm~\ref{alg:1}), then 
$x_v^{\pi}$ will be replaced with $\mathrm{skip}$ 
when calculating \eqref{eqn:actual_probability}.

During the training process, our goal is to maximize the expected total reward $R_{\theta}=\mathbb{E}_{\tau\sim p_{\theta}}\left[w_{x_v,v}\right]$.
Thus, at each training step, given an RL policy  with parameter $\theta$, we sample $n$ action trajectories $\{\tau_i=\{x_{1,i},\cdots, x_{v,i}\}, i\in[n]\}$ and record the corresponding rewards. 
We can get the approximated average reward as $\hat{R}_{\theta}=\frac{1}{n}\sum_{i=1}^n w^i_{x_{i,v},v}$, and
calculate the gradient 
as
\begin{equation} \label{eqn:rl_gradient}
    \nabla_{\theta}\hat{R}_{\theta} = \sum_{i=1}^n\left(\sum_{v\in\mathcal{V}}\nabla_{\theta}\log p_{\theta}(x_{v,i}|I_{u,i})\right)\left(\sum_{v\in\mathcal{V}}w^i_{x_{v,i},v}\right)
\end{equation}
Then, we update the parameter $\theta$ by $\theta=\theta+\alpha\nabla_{\theta}\hat{R}_{\theta}$, where $\alpha$ is the step size. This process repeats until convergence and/or the maximum
number of iterations is reached.

At the beginning of the policy training, we can set a high temperature $t$ to encourage the RL model to explore more aggressively, instead of sticking to the expert's decisions. 
As the 
RL model performance continuously improves, we can reduce the temperature in order to make the RL agent more aware of the downstream 
switching operation. The training process is performed offline
as in the existing RL-based optimizers \citep{L2O_OnlineBipartiteMatching_Toronto_ArXiv_2021_DBLP:journals/corr/abs-2109-10380,L2O_AdversarialOnlineResource_ChuanWu_HKU_TOMPECS_2021_10.1145/3494526}
and described in Algorithm~\ref{alg:training_brief} for one iteration.

\section{Experiment}\label{sec:experiment}

\begin{figure*}[htp]	
	\centering
	\subfigure[Testing with $\rho=0.4$]{
	\includegraphics[width=0.31\textwidth]{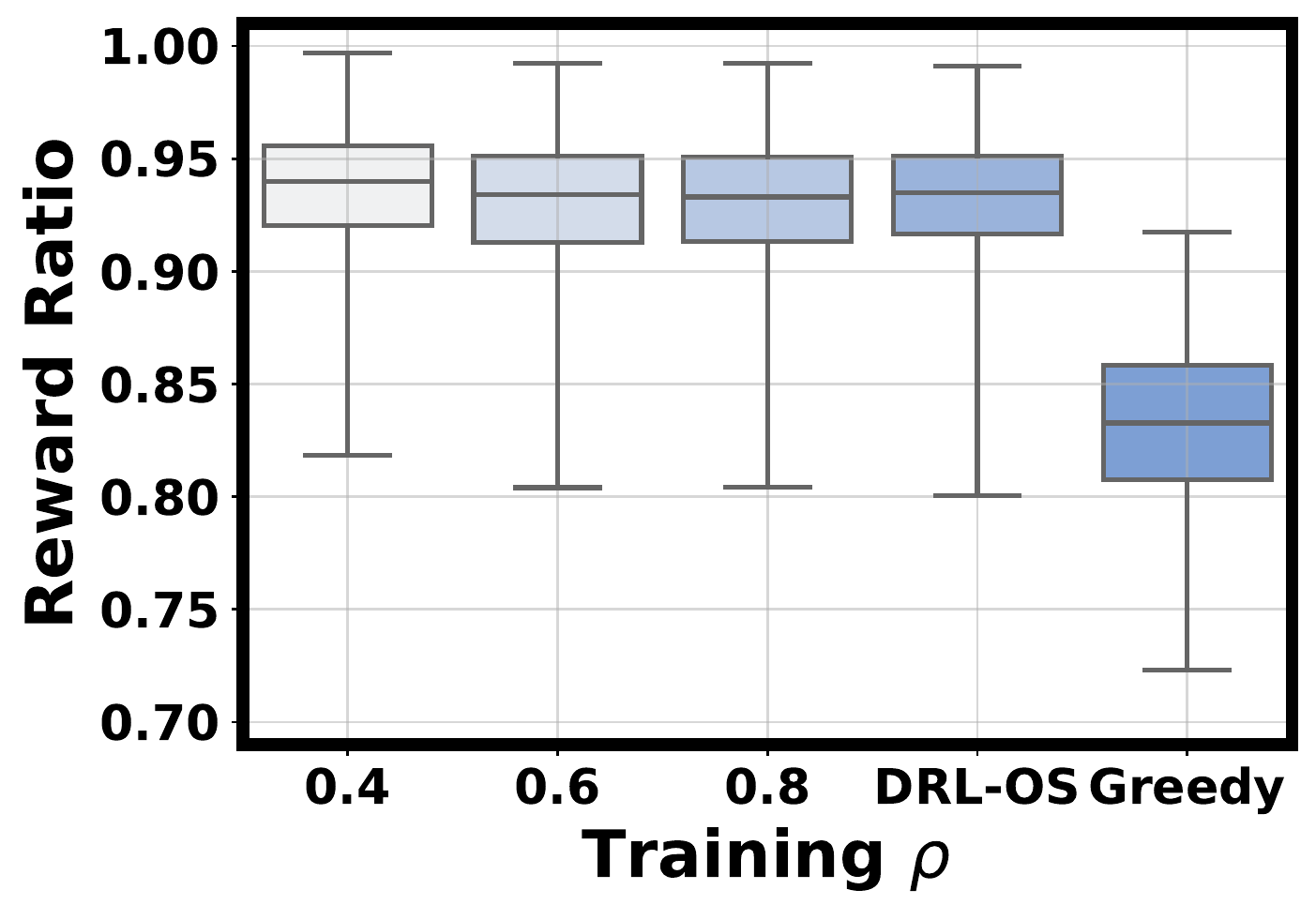}
	}%	
	\subfigure[Testing with $\rho=0.6$]{
	\includegraphics[width=0.31\textwidth]{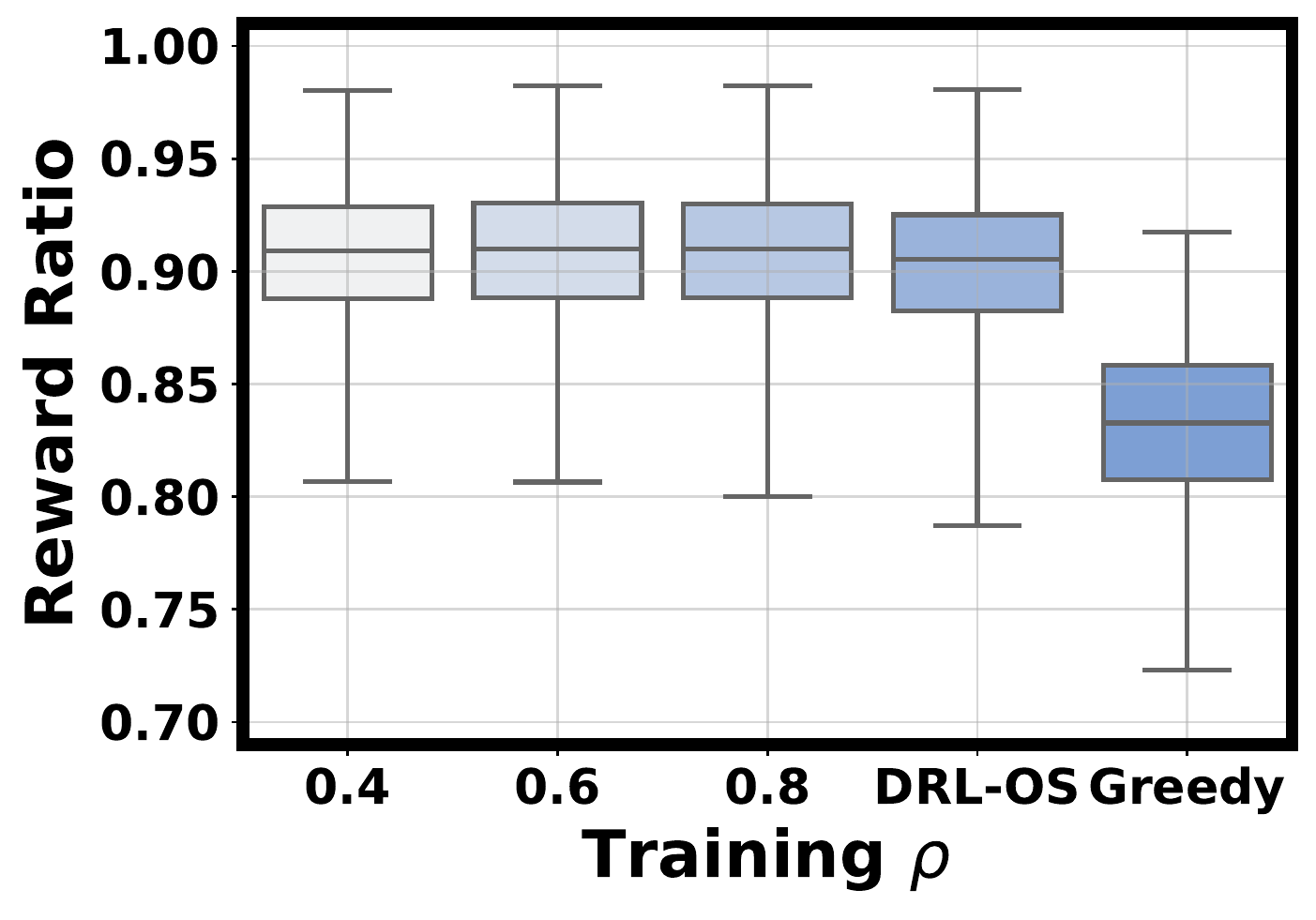}
	}
	\subfigure[Testing with $\rho=0.8$]{
	\includegraphics[width=0.31\textwidth]{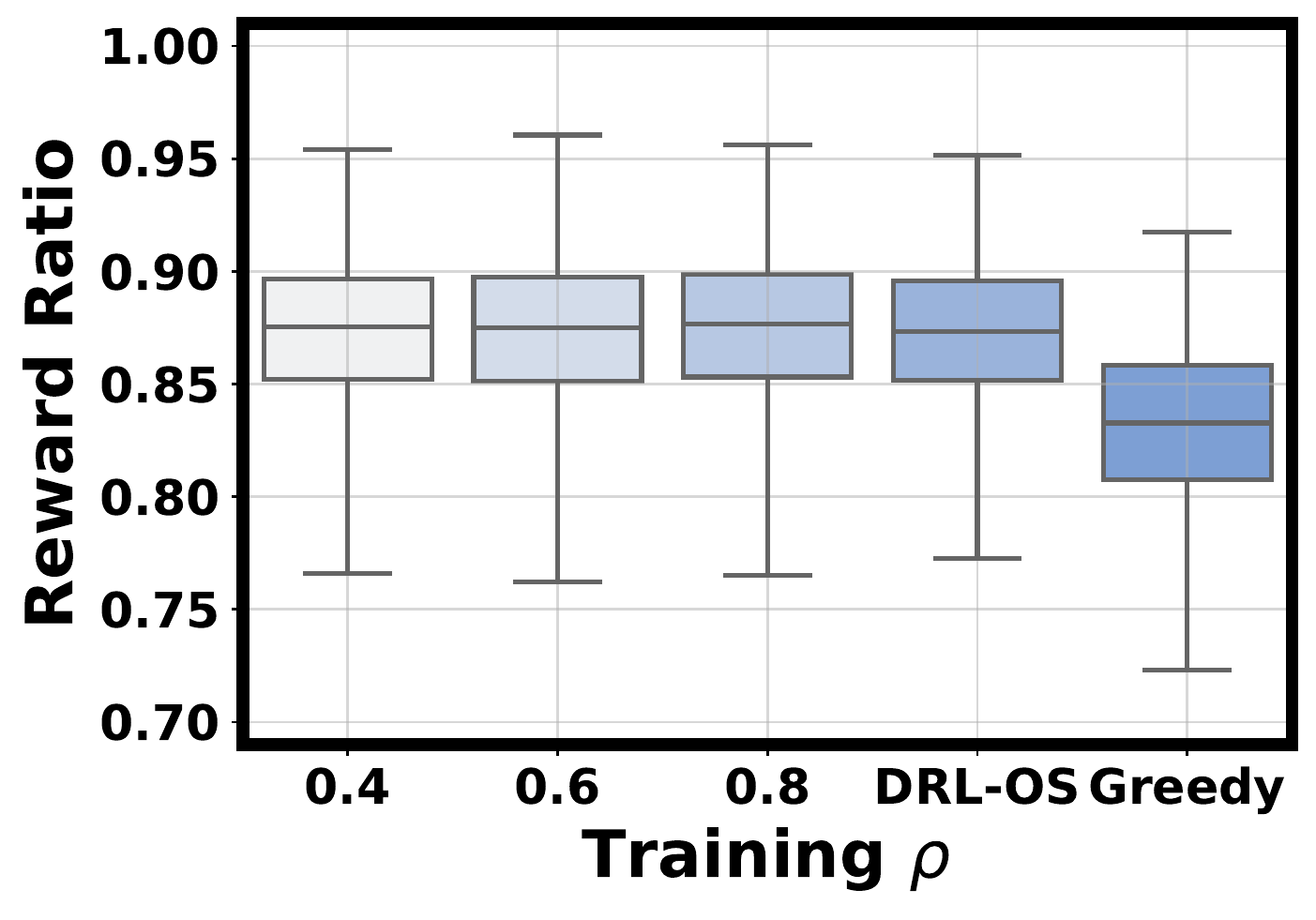}
	}
\vspace{-0.4cm}	
\caption{Boxplot for reward ratio with different $\rho$ within testing dataset.  \greedy and \rob are also shown here for comparison. The best average performance in each figure is achieved by choosing the same $\rho$ during training and testing.}
\label{fig:boxplot}
\end{figure*}

\begin{figure*}[htp!]	
	\centering
	\subfigure[$\rho=0.0$ (i.e., \ml)]{
	\includegraphics[width=0.23\textwidth]{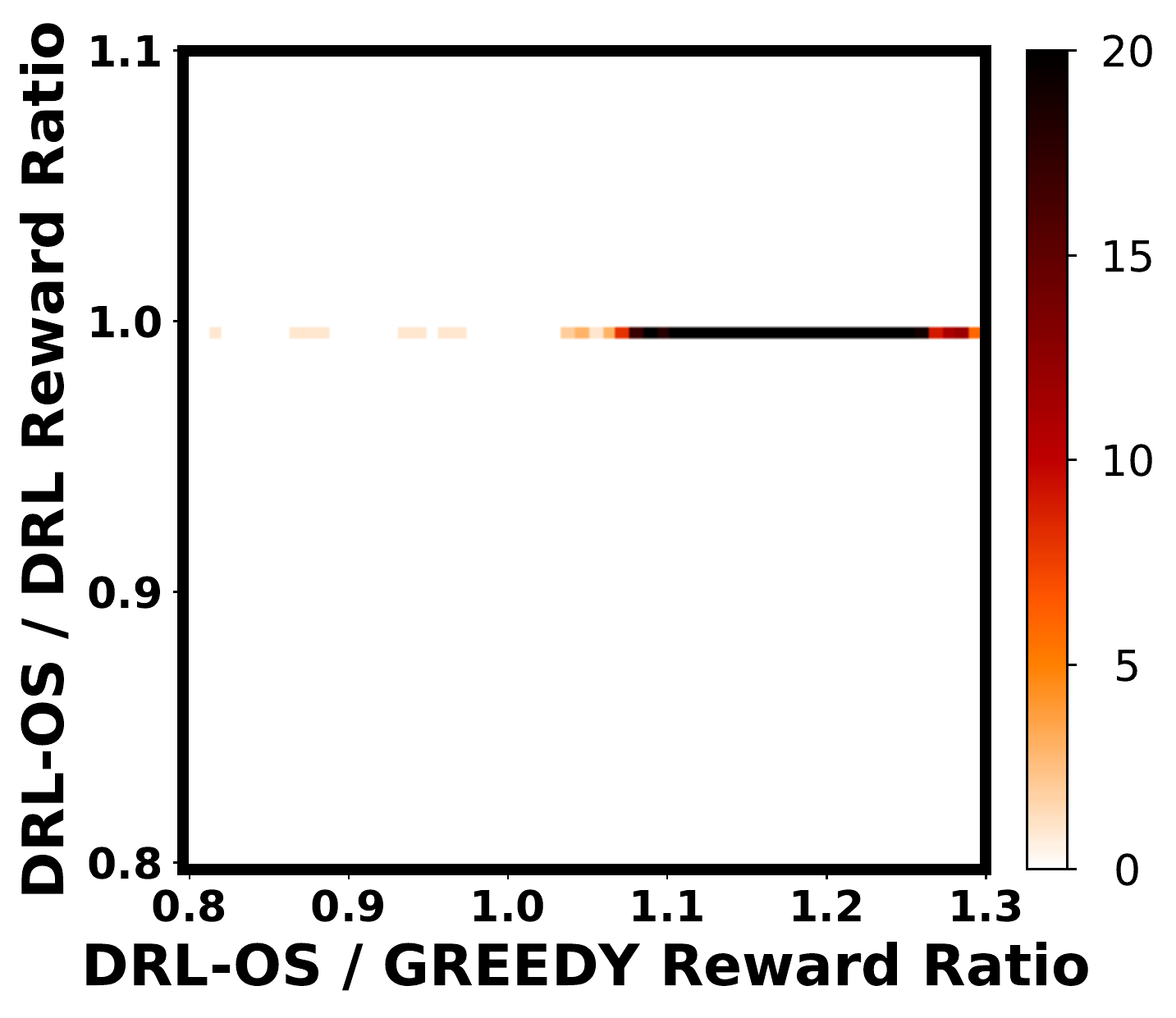}
	}%	
	\subfigure[$\rho=0.4$]{
	\includegraphics[width=0.23\textwidth]{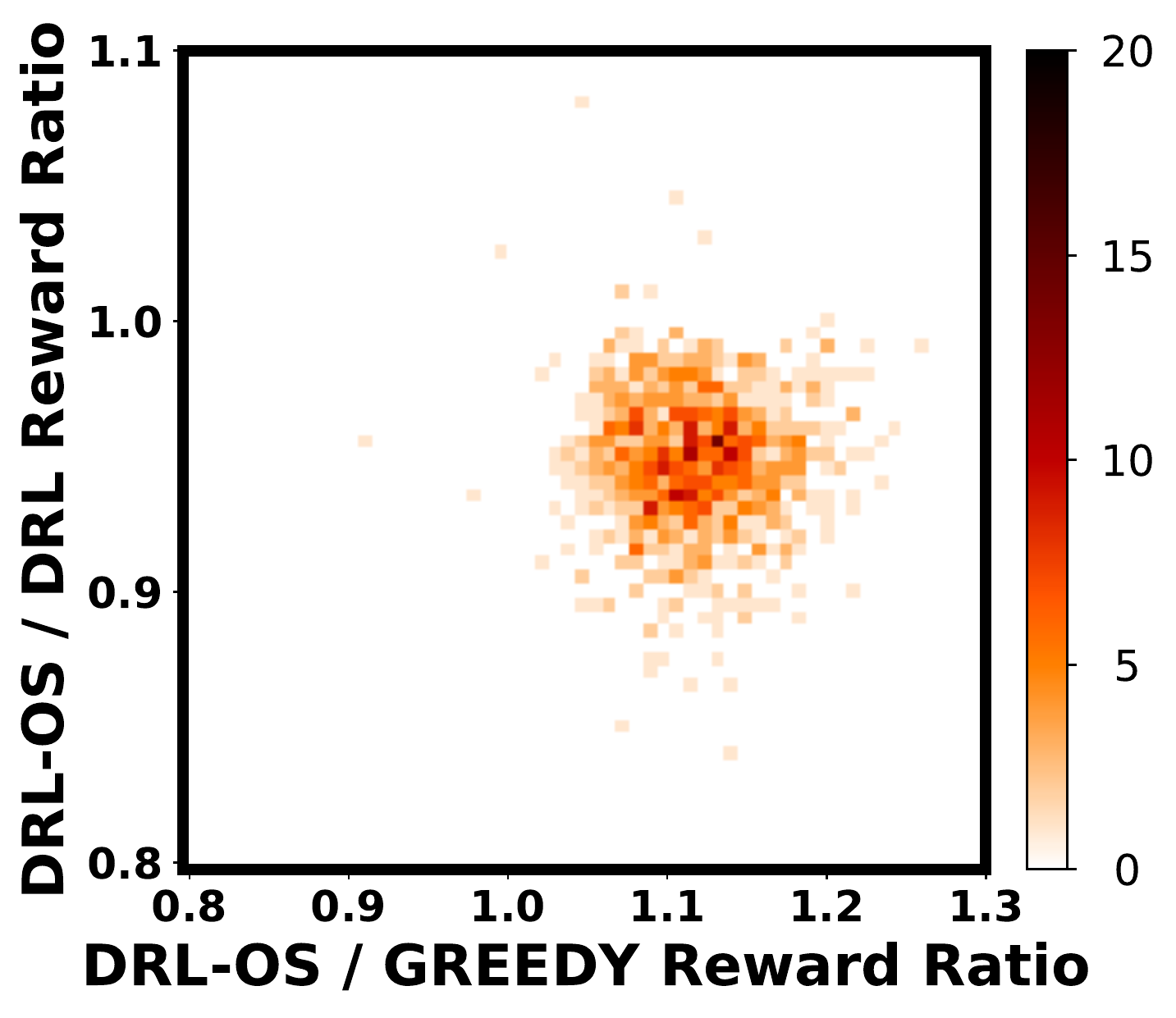}
	}
	\subfigure[$\rho=0.6$]{
	\includegraphics[width=0.23\textwidth]{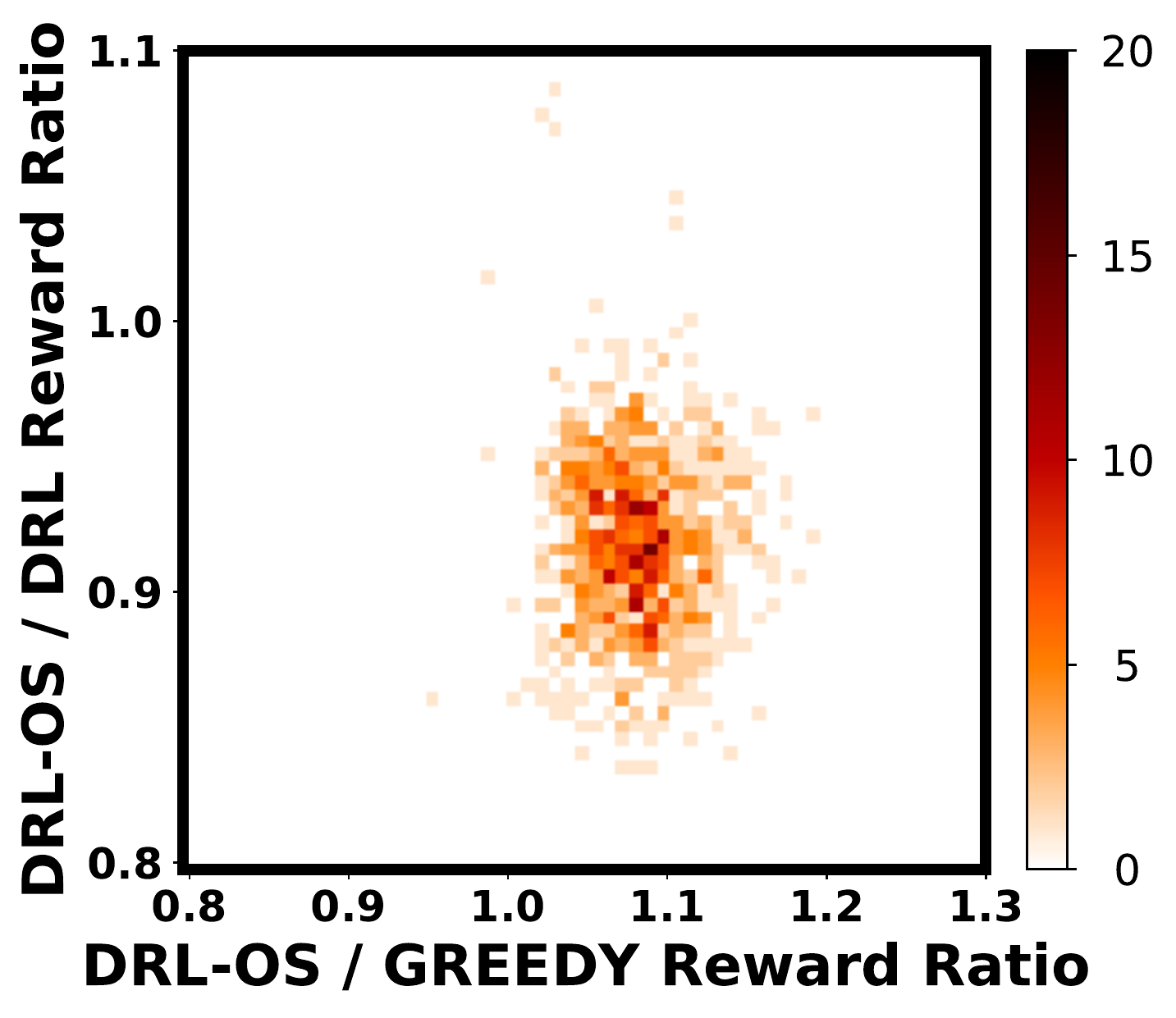}
	}
	\subfigure[$\rho=0.8$ ]{
	\includegraphics[width=0.23\textwidth]{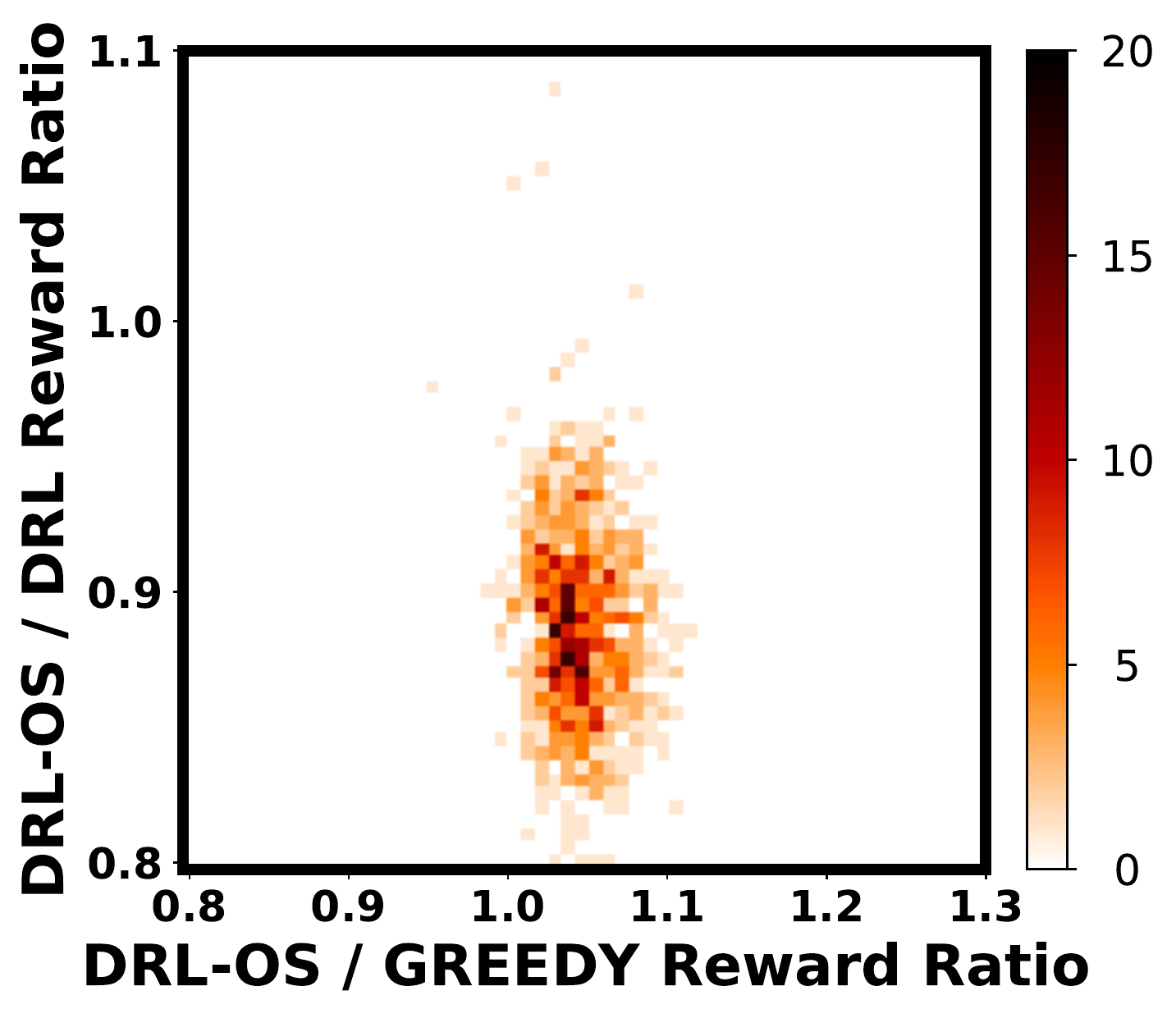}
	}
\vspace{-0.4cm}	
\caption{Histogram of bi-competitive reward ratios of \rob (against
\greedy and \ml) under different $\rho$.}\label{fig:bi-competitiveness}
\end{figure*}

\subsection{Setup}

We conduct experiments based
on the movie recommendation application. Specifically, 
when an user (i.e., online item $v$) arrives, 
we recommend a movie (i.e., offline item $u$) to this user and receive
a reward based on the user-movie preference information. 
We choose the MovieLens  dataset \citep{10.1145/2827872}, which provides a total of 3952 movies, 6040 users and 100209 ratings. We preprocess the dataset to sample movies  and users randomly from the dataset to generate subgraphs, following the same steps as used by \citep{10.1609/aaai.v33i01.33011877} and \citep{L2O_OnlineBipartiteMatching_Toronto_ArXiv_2021_DBLP:journals/corr/abs-2109-10380}. In testing dataset, we  empirically evaluate each algorithm using average reward 
 (\textbf{AVG})  and competitive ratio (\textbf{CR}, against OPT), which represents the average performance and worst case performance, respectively. 
Thus, the value of CR is the empirically worst reward ratio
in the testing dataset.
For fair comparison, all the experimental settings like capacity $c_u$ follow
those used in \citep{L2O_OnlineBipartiteMatching_Toronto_ArXiv_2021_DBLP:journals/corr/abs-2109-10380}. 
More details about the setup and training  are in Appendix~\ref{appendix_experiments}.   

\textbf{Baseline Algorithms.} We consider the following baselines.
All the RL policies are trained offline with
the same architecture and applicable hyperparameters.
\textbf{\opt:} The offline optimal oracle has the complete information about the bipartite graph. We use the Gurobi optimizer to find the optimal offline solution.
\textbf{\greedy:} At each step, \greedy  selects the available offline item with highest weight. 
 \textbf{\ml:} It uses the same architecture as in \ouralg, but does not consider online switching for training or inference. That is,
    the RL model is both trained and tested with $\rho=0$.
    More specifically, our RL architecture has 3 fully connected layers, each with 100 hidden nodes.
\textbf{\rob (DRL-OnlineSwitching):} We apply online switching
    to the same RL policy used by \ml during inference. That is,
    the RL model is trained with $\rho=0$, but tested with a different
    $\rho>0$. This is essentially an existing ML-augmented algorithm that
    uses the standard practice (i.e., pre-train a standalone RL policy) \cite{SOCO_ML_ChasingConvexBodiesFunction_Adam_COLT_2022}.

Our baselines include all those considered in \citep{L2O_OnlineBipartiteMatching_Toronto_ArXiv_2021_DBLP:journals/corr/abs-2109-10380}. 
%Although \ouralg is not limited to any specific expert algorithm,
In the no-free-disposal setting, the best competitive
ratio is 0 in general adversarial cases \citep{OnlineMatching_Booklet_OnlineMatchingAdAllocation_Mehta_2013_41870}.
Here, we use \greedy as the expert, because
the recent study \citep{L2O_OnlineBipartiteMatching_Toronto_ArXiv_2021_DBLP:journals/corr/abs-2109-10380} has shown that 
 \greedy performs better than other alternatives and is a strong baseline.

\subsection{Results}

\textbf{Reward comparison.} We compare \ouralg with baseline algorithms
in Table~\ref{table:rho}. First, we see that 
\ml has the highest average reward, but its empirical competitive ratio is the lowest. The expert algorithm \greedy is fairly robust, but has a lower average award than RL-based policies. 
Second, \rob can improve the competitive ratio compared to \ml. But, its RL policy is trained alone without being aware of the online switching.
Thus, by making the RL policy aware
of online switching, \ouralg can  improve the average reward compared to \rob. 
Specifically, by training \ouralg 
using the same $\rho$ as testing it,
we can obtain both the highest average cost and the highest competitive ratio.
One exception is the minor decrease of competitive ratio when $\rho=0.8$ for testing. 
This is likely due to the dataset 
and a few hard instances can affect the empirical competitive ratio, which also explains why the empirical competitive ratio is not necessarily monotonically increasing in the $\rho\in[0,1]$. Nonetheless,
unlike \ml that may only 
work well empirically without guarantees, 
\ouralg offers provable robustness while exploiting
the power of RL to improve the average performance.
The boxplots in Fig.~\ref{fig:boxplot} visualizes the reward ratio distribution of \ouralg, further validating the importance of switching-aware training.

\textbf{Impact of $\rho$.} To show the impact of $\rho$, we calculate 
the bi-competitive reward ratios. Specifically, for each problem instance, the bi-competitive ratio compares the actual reward against those
of \greedy and RL model, respectively. To highlight the effect of online switching, we focus on  \rob (i.e., training the RL with $\rho=0$) whose training process of RL model is not affected by $\rho$, because the RL model trained 
with $\rho>0$ in \ouralg does not necessarily perform well on its own and the reward ratio
of \ouralg to its RL model is not meaningful.
The histogram of the bi-competitive ratios are visualized in Fig.~\ref{fig:bi-competitiveness}. When $\rho=0$, the ratio of \rob / \ml is always 1 unsurprisingly, since \rob are essentially the same as \ml in this case
(i.e., both trained and tested with $\rho=0$). 
 With a large $\rho$ (e.g. 0.8) for testing, the reward ratios of \rob/\greedy  for most samples are around 1, which means the robustness is achieved, as proven by our theoretical analysis. But on the other hand, \rob has limited flexibility and can less exploit the good average performance of \ml. Thus,
 the hyperparameter $\rho \in [0,1]$ governs the tradeoff between average performance and robustness relative to the expert and, like other hyperparameters, can be tuned 
 to maximize the average performance subject to the robustness requirement.

We also consider a crowdsourcing application, as provided by the gMission dataset \citep{chen2014gmission}. Additional results for gMission are deferred to Appendix~\ref{appendix_experiments}.

\section{Conclusion}

 In this paper, we propose \ouralg
for edge-weighted online bipartite matching. 
\ouralg
includes a novel online switching operation to decide whether to follow
the expert's decision or the RL decision for each online item arrival.
We prove that for any $\rho\in[0,1]$, \ouralg is $\rho$-competitive against any expert online algorithms, which
directly translates a bounded competitive ratio against \opt if the expert algorithm itself has one. 
We also train
the RL policy by explicitly considering the online switching operation so as to improve the average performance. Finally, we run
empirical experiments to validate \ouralg.

There are also interesting problems that remain open, such as
how to incorporate multiple RL models or experts and what
 the performance bound of \ouralg is compared to pure RL in terms
of the average reward.

\section*{Acknowledgement}

This work was supported in part by the U.S. National Science Foundation under
the grant CNS-1910208.

\bibliography{main_lomar.bbl}
\bibliographystyle{icml2023}

\onecolumn
\newpage
\appendix

\section*{Appendix}
In the appendix, we show 
the experimental setup and additional results (Appendix~\ref{appendix_experiments}),
algorithm details for the free-disposal setting (Appendix~\ref{appendix_free_disposal}),
and finally the proof of Theorem~\ref{thm:cr} (Appendix~\ref{appendix_proof}).

\section{Experimental Settings and Additional Results}\label{appendix_experiments}

Our implementation of all the considered algorithms, including \ouralg, is based on the source codes provided by \cite{L2O_OnlineBipartiteMatching_Toronto_ArXiv_2021_DBLP:journals/corr/abs-2109-10380}, which includes codes for training the RL model, data pre-proposing and performance evaluation. 
We conduct experiments
on two real-world datasets: MovieLens  \cite{10.1145/2827872}
and gMission \cite{chen2014gmission}.

\subsection{MovieLens}

\subsubsection{Setup and Training}\label{appendix_setup_movielens}
We first sample $u_0$ movies from the original MovieLens dataset \cite{10.1145/2827872}. We then sample $v_0$ users and make sure each user can get at least one movie;
otherwise, we remove the  users 
that have no matched movies,
and resample new users. After getting the topology graph, we use Gurobi 
to find  the optimal matching decision.  
In our experiment, we set $u_0 = 10$ and $v_0 = 60$ to generate the training and testing datasets. The number of graph instances in the training and testing datasets are 20000 and 1000, respectively.
For the sake of reproducibility and fair comparision, our settings  follows the same setup of \cite{L2O_OnlineBipartiteMatching_Toronto_ArXiv_2021_DBLP:journals/corr/abs-2109-10380}. In particular,
the general movie recommendation problem  belongs to online submodular optimization, but it can actually be equivalently mapped to edge-weighted online bipartite matching with no free disposal under the setting considered in \cite{L2O_OnlineBipartiteMatching_Toronto_ArXiv_2021_DBLP:journals/corr/abs-2109-10380}. So by default, the capacity $c_u$ for each offline node is set as 1 and $w_{u, \max} = 5$. While \ouralg can use any RL architecture,
we follow the design of \emph{inv-ff-hist} proposed by \cite{L2O_OnlineBipartiteMatching_Toronto_ArXiv_2021_DBLP:journals/corr/abs-2109-10380}, which empirically demonstrates  the best performance among all the considered architectures. 

The input to our considered RL model is the edge weights $w_{uv}$ revealed by the online items plus some historical information,
which includes:
 Mean and variances of each offline node's weights;
 Average degree of each offline nodes; Normalized step size;
 Percentage of offline nodes connected to the current node; 
 Statistical information of these already matched nodes' weights (maximum, minimum, mean and variance);
Ratio of matched offline node; 
Ratio of skips up to now;
Normalized reward with respect to the offline node number. 
For more details of the historical information, readers are referred to  Table~1 in \cite{L2O_OnlineBipartiteMatching_Toronto_ArXiv_2021_DBLP:journals/corr/abs-2109-10380}.

\begin{wrapfigure}[]{r}{0.42\textwidth}
    \vspace{-0.4cm}
    \includegraphics[trim=0 0cm 0 0, clip, width=0.4\textwidth]{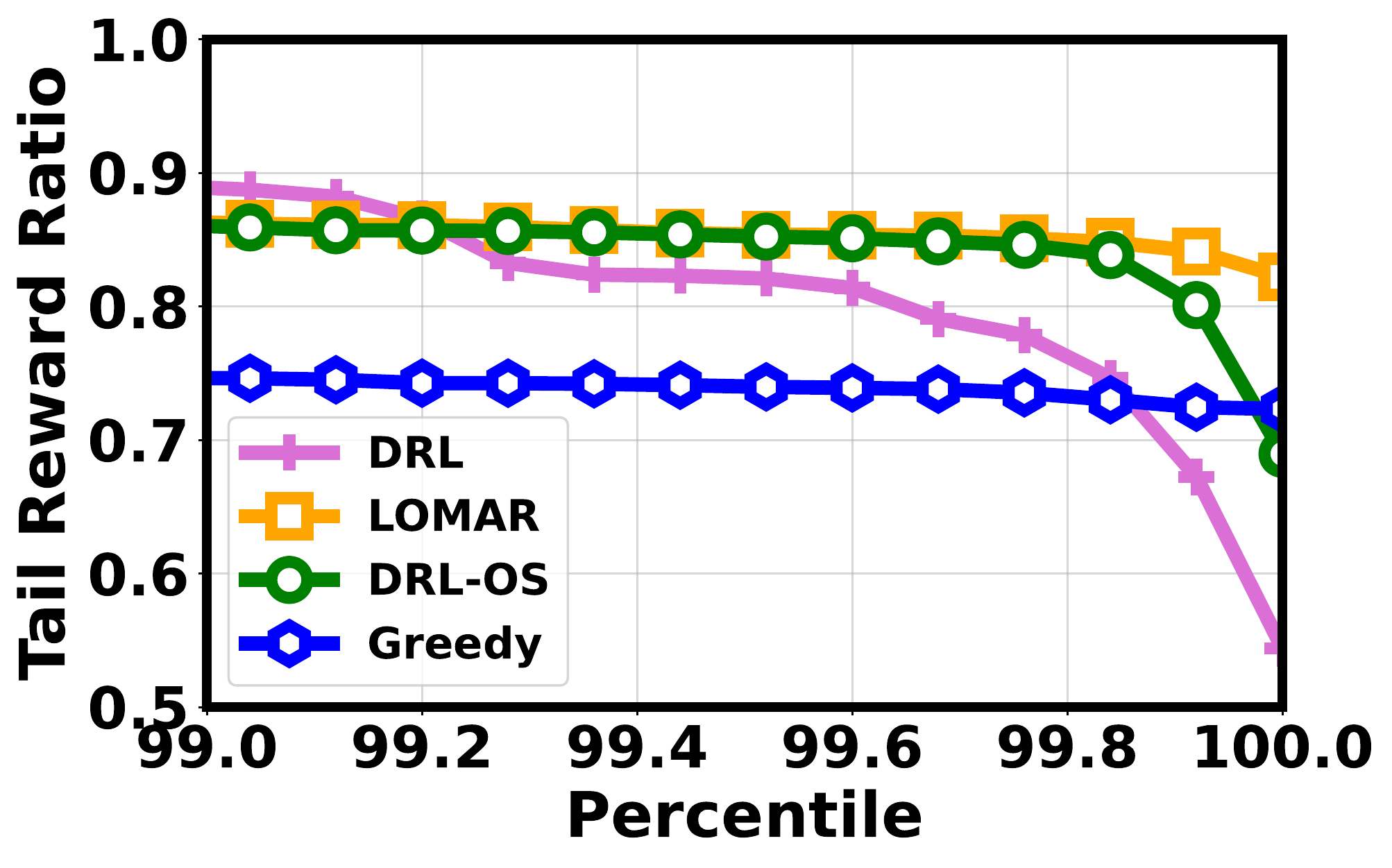}
    \vspace{-0.5cm}
    \caption{Tail reward ratio comparison. In this experiment, we set $\rho = 0.4$ for DRL-OS and \ouralg.}
    
    \label{fig:tail_cost}
\end{wrapfigure}

For applicable algorithms
(i.e., \ml, \rob, and \ouralg),
we train the RL model for 300 epochs in the training dataset with a batch size of 100. 
In \ouralg, the  parameter $B=0$ is
used to follow the strict definition of competitive ratio.
We test the algorithms on the testing dataset to obtain the average reward and the worst-case competitive ratio empirically.  
By setting $\rho = 0$ for training, \ouralg is equivalent to the vanilla inv-ff-hist RL model (i.e., \ml) used in \cite{L2O_OnlineBipartiteMatching_Toronto_ArXiv_2021_DBLP:journals/corr/abs-2109-10380}. 
Using the same problem setup, we can reproduce the same results shown in \cite{L2O_OnlineBipartiteMatching_Toronto_ArXiv_2021_DBLP:journals/corr/abs-2109-10380}, which reaffirms the correctness of our data generation  and training process.

Additionally, training the RL model in \ouralg usually takes less than 8 hours on a shared research cluster with one NVIDIA K80 GPU, which is almost the same as the training the  model for \ml in a standalone manner (i.e., setting $\rho=0$ without considering online switching).

\subsubsection{Additional Results}

In Table~\ref{table:rho}, we have empirically demonstrated that 
\ouralg achieves the best tradeoff
between
the average reward and competitive ratio.
In Fig ~\ref{fig:tail_cost}, we further demonstrate that \ouralg not only achieves a better worst-case competitive ratio (at 100.0\%). The tail reward ratio  of \ouralg
 is also good compared to the baseline algorithms.
Specifically, we show the percentile
of reward ratios (compared to the optimal offline algorithm) ---
the 100\% means the worst-case
empirical reward ratio (i.e., competitive ratio). We see
that \ml has a bad high-percentile reward ratio
and lacks performance robustness, although its lower-percentile cost ratio
is better. This is consistent with the good average performance
of \ouralg. Because
of online switching, both \rob and \ouralg achieve better robustness,
and \ouralg is even better due to its awareness of the online switching operation during its training process.
The expert \greedy has a fairly stable competitive ratio, showing its good robustness. But, it can be outperformed by other algorithms when we look at lower-percentile reward ratio. 

\subsubsection{Results for another expert algorithm}

\begin{figure*}[htp]	
	\centering
	\subfigure[Reward ratio against \opt]{
	\includegraphics[width=0.31\textwidth]{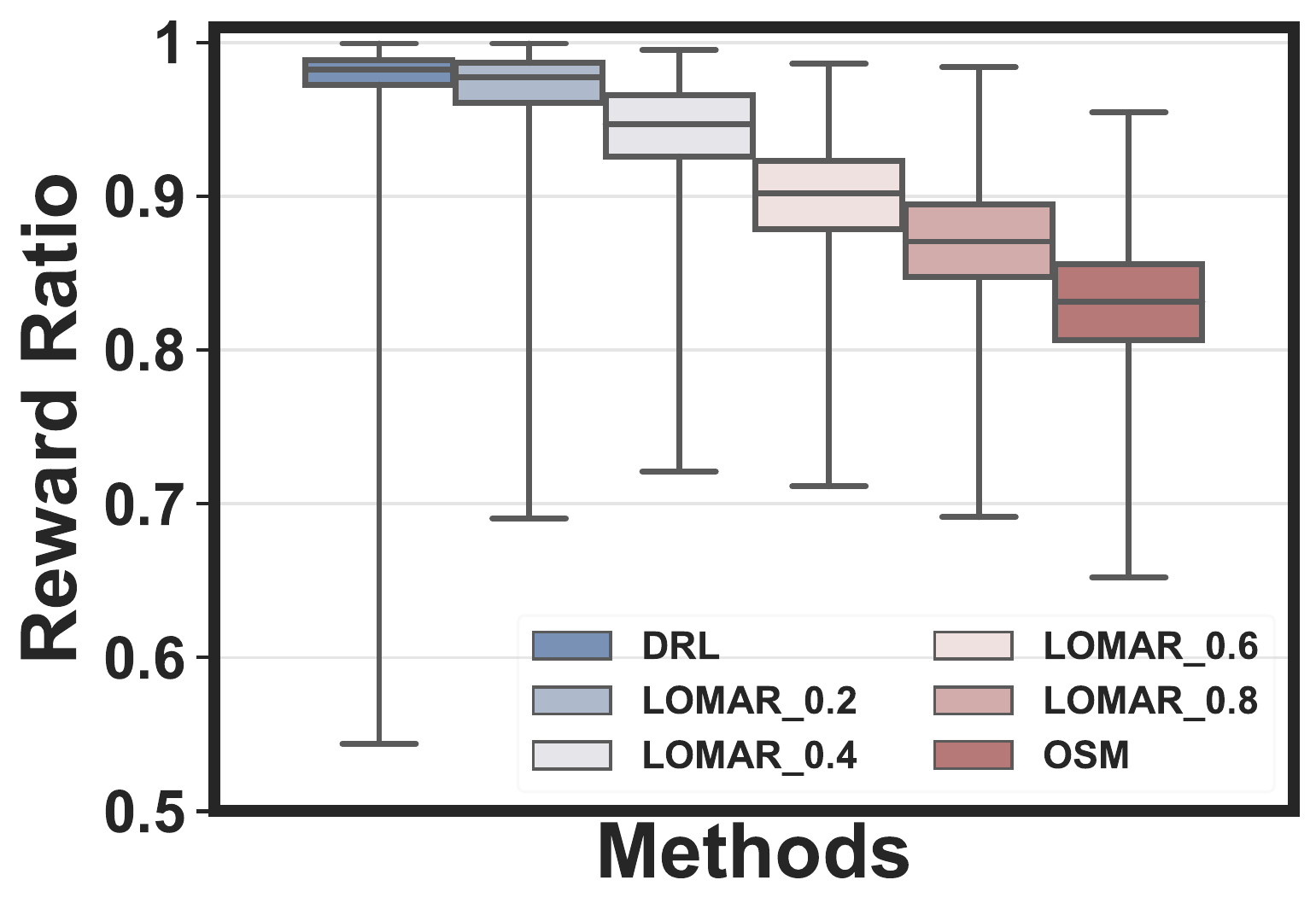}
	}\quad\quad\quad
	\subfigure[Reward ratio against \osm]{
	\includegraphics[width=0.31\textwidth]{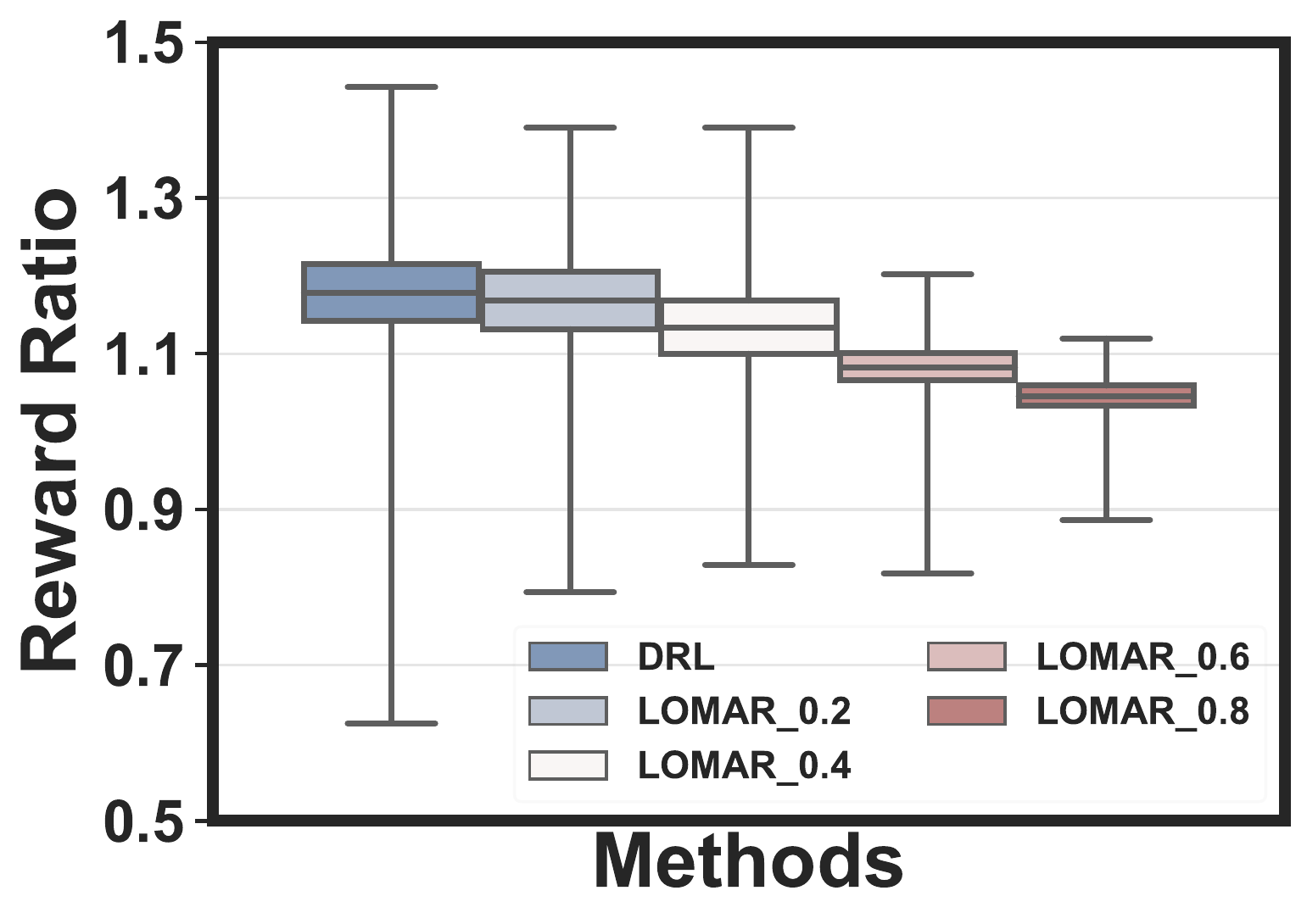}
	}
\vspace{-0.2cm}	
\caption{Reward ratio distribution (\osm as the expert)}
\label{fig:cost_ratio_osm}
\end{figure*}

Optimally competitive expert algorithms 
have been developed 
under the assumptions of random oder and/or i.i.d. rewards
of different online items. In particular,
by considering the random order setting, 
\osm (online secretary matching) has the optimal competitive
ratio of $1/e$  \citep{kesselheim2013optimal}. Note that
the competitive ratio for \osm is average over the random order of online items, while the rewards can be adversarially chosen. 
We show the empirical results in Fig.~\ref{fig:cost_ratio_osm}.
As  \osm skips the first $|\mathcal{V}|/e$ online items, it actually
does not perform (in terms of the empirical worst-case cost ratio) as well as the default expert \greedy in our experiments
despite its guaranteed competitive ratio against \opt.  
That said, we still observe the same trend as using \greedy for
the expert: by tuning $\rho\in[0,1]$, \ouralg achieves
a good average performance while guaranteeing the competitiveness
against the expert \osm (and against \opt as \osm itself is optimally competitive against \opt).

Fig.~\ref{fig:cost_ratio_osm} shows the empirical results in our testing dataset, 
which does not strictly satisfy the random order assumption required by \osm.
Next, to satisfy the random order assumption, we select a typical problem instance
and randomly vary the arrival orders of online items.
We show the cost ratio averaged over the random arrival order in
Table~\ref{table:osm}. Specifically,
we calculate each cost ratio by 100 different random orders, and 
repeat this process 100 times. We show
the mean and stand deviation of the average cost ratios
(each averaged over 100 different random orders). We
see that \ouralg improves the average cost ratio compared to \osm 
under the random order assumption. While \ml has a better average
cost for this particular instance, it does not provide any guaranteed
worst-case robustness as \ouralg.

\begin{table}[htp]
\centering
\small
\begin{tabular}{c|c|c|c|c|c|c} 
\toprule
     & \ml & \ouralg $\rho=0.2$ & \ouralg $\rho=0.4$ & \ouralg $\rho=0.6$ & \ouralg $\rho=0.8$ & OSM \\     \hline
Mean & 0.9794  & 0.9688                    & 0.9431                    & 0.9095                    & 0.8799                    & 0.8459  \\ \hline
Std  & 0.0074  & 0.0082                    & 0.0078                    & 0.0086                    & 0.0084                    & 0.0084  \\
\bottomrule
\end{tabular}
\caption{Reward ratio (averaged over the random arrival order)
for a typical graph instance}\label{table:osm}
\end{table}

\subsubsection{Results for the free-disposal setting}

\begin{figure*}[!t]	
	\centering
	\subfigure[Reward ratio against \opt]{
	\includegraphics[width=0.31\textwidth]{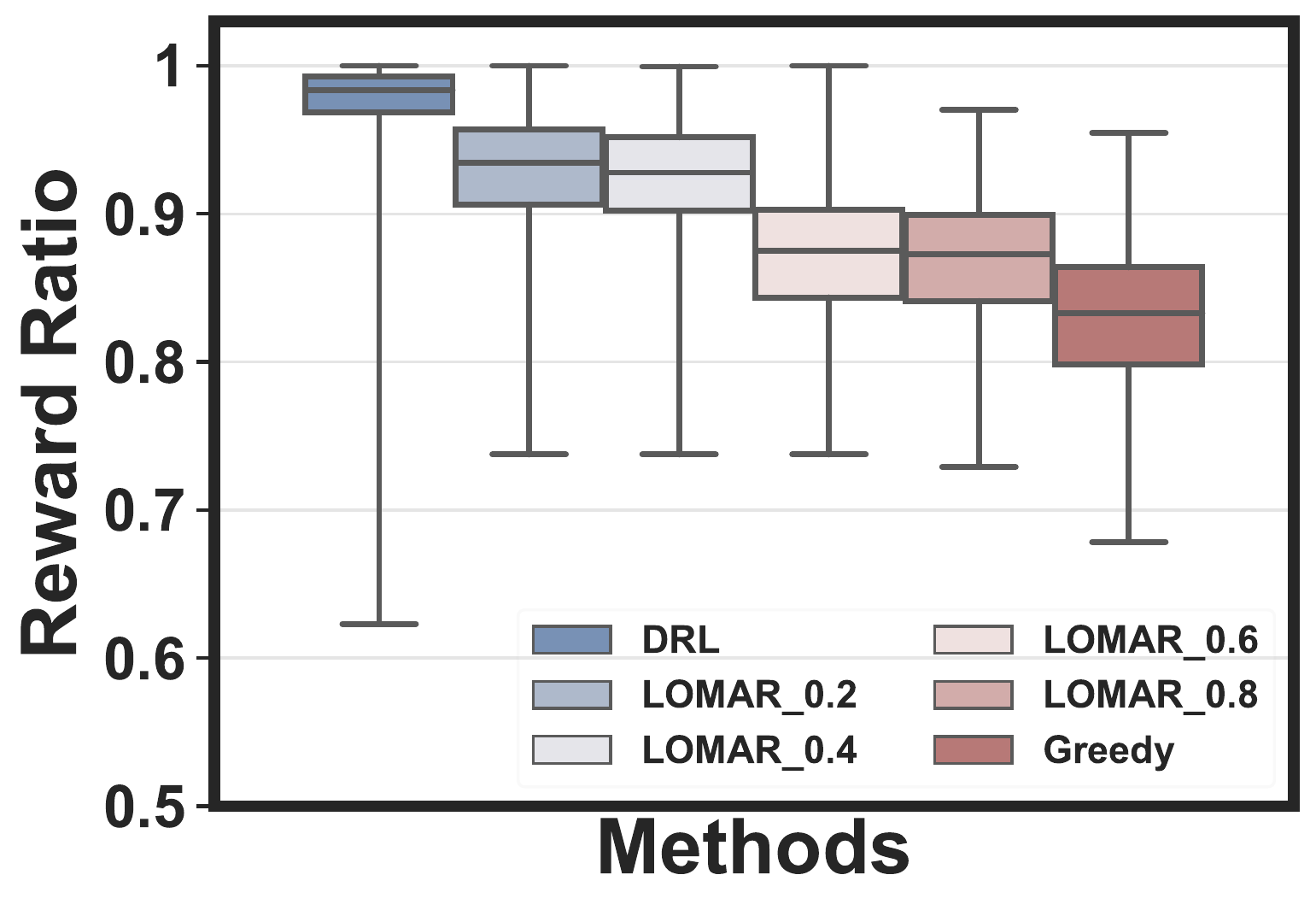}
	}\quad\quad\quad
	\subfigure[Total reward]{
	\includegraphics[width=0.31\textwidth]{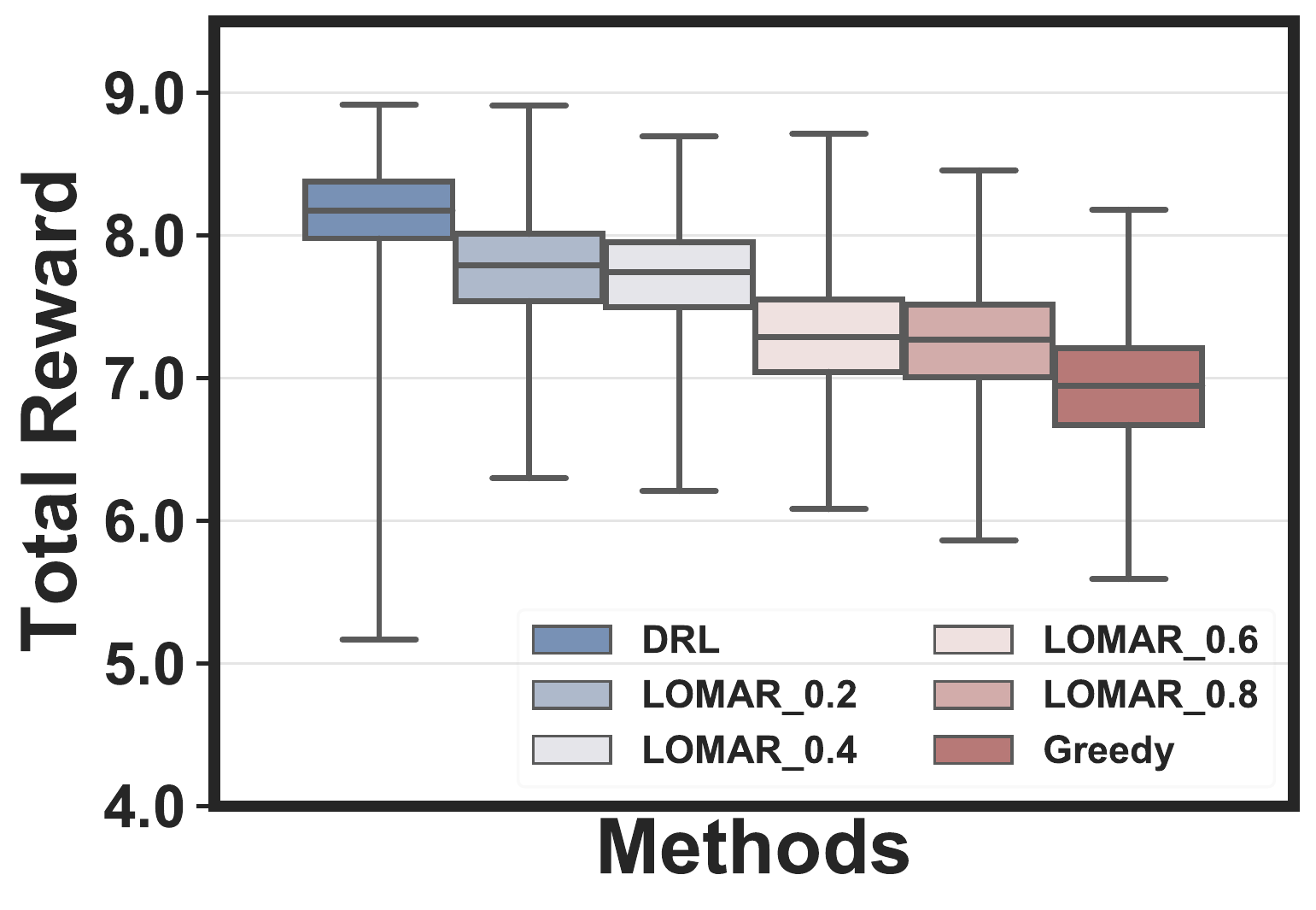}
	}
\vspace{-0.2cm}	
\caption{Reward ratio and total reward distributions for the free-disposal
setting (\greedy as the expert).}
\label{fig:boxplot_free_disposal}
\end{figure*}

For the free-disposal setting, we use the same parameter (e.g. RL architecture, learning rates) and datasets as in the no-free-disposal case. 
By modifying the implementation of the public codes released by \cite{L2O_OnlineBipartiteMatching_Toronto_ArXiv_2021_DBLP:journals/corr/abs-2109-10380} that focus on no-free-disposal matching, 
we consider a 5 $\times$ 60 graph and allow each offline node
to be matched with multiple online items, while only the maximum reward
for each offline node is considered.
In Table~\ref{table:free_dis} and Fig.~\ref{fig:boxplot_free_disposal}, we use \greedy as the expert and evaluate \ouralg with different $\rho$ parameters. The empirical results show a similar trend as our experiments under the no-free-disposal setting: with
a smaller $\rho$, the average performance of \ouralg is closer to \ml. 

\begin{table}[!t]
\centering
\small
\begin{tabular}{c|c|c|c|c|c|c} 
\toprule
     & \ml &\ouralg $\rho=0.2$  & \ouralg $\rho=0.4$ & \ouralg $\rho=0.6$ & \ouralg $\rho=0.8$ & \greedy \\     
     \hline
AVG  &8.172 &7.764 & 7.712 & 7.298 & 7.256 & 6.932\\ 
\hline
CR  &0.623  &0.738  & 0.738 & 0.738 & 0.729 & 0.678 \\
\bottomrule
\end{tabular}
\caption{Average reward and competitive ratio comparison between different algorithms. The average reward of \opt is 8.359.}
\label{table:free_dis}
\end{table}

\subsection{gMission}

The gMission dataset \cite{chen2014gmission}  considers a crowdsourcing application, where the goal is to assign the tasks (online items) to workers (offline items). The edge weight between a certain online task and each worker can be calculated by the product of the task reward and the worker's success probability, which is determined by the physical location of workers and the type of tasks. 
Our goal is to maximize the total  reward given the capacity of each worker, which perfectly fits into our formulation in Eqn.~\eqref{eqn:offline_problem}.

We use the same data processing and RL architecture design as introduced in Section~\ref{appendix_setup_movielens}. We  train \ouralg with different $\rho$ in the gMission dataset by setting $u_0 = 10$, $v_0 = 60$, $w_{u,\max} = 1$.
Again, we use \greedy as the expert,
which is an empirically strong baseline algorithm as shown in \cite{L2O_OnlineBipartiteMatching_Toronto_ArXiv_2021_DBLP:journals/corr/abs-2109-10380}. Our results are all consistent with those presented in \cite{L2O_OnlineBipartiteMatching_Toronto_ArXiv_2021_DBLP:journals/corr/abs-2109-10380}.

\subsubsection{Testing on $10 \times 60$}

In our the first result, we generate a testing  dataset with $u_0 = 10$ and $v_0 = 60$, which is the same setting as our training dataset. In other words,
the training and testing datasets have similar distributions. Specifically,
\greedy's average reward and competitive ratio are 4.508 and 0.432, while  these two values for \ml are 5.819 and 0.604, respectively.
Thus, \ml performs outperforms \greedy in both average performance and the worst-case performance.

\begin{table}[htp!]
\centering
\small
\begin{tabular}{c|cc|cc|cc|cc|cc} 
\toprule
\multicolumn{1}{c}{} & \multicolumn{2}{c}{\rob} & \multicolumn{2}{c}{\ouralg $\rho=0.4$} & \multicolumn{2}{c}{\ouralg $\rho=0.6$} & \multicolumn{2}{c}{\ouralg $\rho=0.8$} & \multicolumn{2}{c}{\ouralg $\rho=0.9$}  \\ 
\hline
$\rho$ in Testing  & AVG   & CR    & AVG   & CR     & AVG   & CR     & AVG   & CR     & AVG   & CR \\ 
\hline
0.4     & 5.553 & \textbf{0.599}     & \textbf{5.573} & 0.598  & 5.553 & 0.598  & 5.523 & 0.598  & 5.535 & 0.598   \\ 
\hline
0.6     & 5.389 & 0.591 & \textbf{5.429} & \textbf{0.619} & 5.420 & 0.619  & 5.403 & 0.623  & 5.402 & 0.623   \\ 
\hline
0.8     & 5.102 & 0.543 & \textbf{5.115} & \textbf{0.543} & 5.111 & 0.523  & 5.110 & 0.521  & 5.107 & 0.521   \\ 
\hline
0.9     & 4.836 & 0.495 & 4.836 & 0.495  & 4.839 & 0.495  & 4.839 & 0.540  & \textbf{4.839} & \textbf{0.540}  \\
\bottomrule
\end{tabular}
\vspace{0.2cm}
\caption{Performance comparison in gMission $10\times60$ for different $\rho$. \ouralg with $\rho=y$ means \ouralg is trained with $\rho=y$.}
\label{table:gmission_10by60}
\end{table}
Next, we show the results for \ouralg and \rob under different $\rho\in[0,1]$ in
Table~\ref{table:gmission_10by60}.
In general, by setting a larger $\rho$ for inference, both \ouralg and \rob are closer to the expert algorithm \greedy, because there is less freedom for the RL decisions. 
As a result, when $\rho$ increases during inference, the average rewards of both
\rob and \ouralg decrease, although they have guaranteed robustness whereas \ml does not.
Moreover, by training the RL model with explicit awareness of online switching, \ouralg can have a higher average cost than \rob, which reconfirms the benefits of training the RL model by considering its downstream operation. Interestingly,
by setting $\rho$ identical
for both training and testing, the average reward may not always be the highest
for \ouralg. This is partially because
of the empirical testing dataset.
Another reason is that, in this test, \ml alone already performs the best (both on average and in the worst case). Hence, by setting a smaller $\rho$ for inference,
\ouralg  works better empirically though it is trained under a different $\rho$. Nonetheless, this does not void the benefits of guaranteed robustness in \ouralg. The empirically better performance of \ml lacks guarantees, which we show as follows.

\subsubsection{Testing on $100\times100$}

In our second test, we consider an opposite case compared to the first one. We generate a testing  dataset with $u_0 = 100$ and $v_0 = 100$, which is different from the training dataset setting.
As a result, the training and testing datasets have very different distributions, making \ml perform very badly. Specifically,
\greedy's average reward and competitive ratio are 40.830 and 0.824, and  these two values for \ml are 32.938 and 0.576, respectively.  \ml has an even lower average reward than \greedy, showing its lack of performance robustness.

\begin{table}[htp!]
\centering
\small
\begin{tabular}{c|cc|cc|cc|cc|cc} 
\toprule
\multicolumn{1}{c}{} & \multicolumn{2}{c}{\rob} & \multicolumn{2}{c}{\ouralg $\rho=0.4$} & \multicolumn{2}{c}{\ouralg $\rho=0.6$} & \multicolumn{2}{c}{\ouralg $\rho=0.8$} & \multicolumn{2}{c}{\ouralg $\rho=0.9$}  \\ 
\hline
$\rho$ in Testing    & AVG    & CR      & AVG    & CR       & AVG   & CR       & AVG    & CR       & AVG   & CR        \\ 
\hline
0.4        & 33.580 & 0.604   & 37.030 & 0.707    & 38.199  & 0.750    & 38.324 & 0.750    & \textbf{38.538} & \textbf{0.766}      \\ 
\hline
0.6        & 34.973 & 0.680   & 37.490 & 0.731    & 38.518  & 0.762    & 38.505 & 0.756    & \textbf{38.727} & \textbf{0.767}      \\ 
\hline
0.8        & 37.939 & 0.758   & 38.866 & 0.775    & 39.502  & 0.782    & 39.385 & \textbf{0.794}    & \textbf{39.552} & 0.781     \\ 
\hline
0.9        & 39.772 & 0.794   & 40.057 & 0.803    & \textbf{40.377} & 0.806    & 40.239 & \textbf{0.812}    & 40.332  & 0.798     \\
\bottomrule
\end{tabular}
\vspace{0.2cm}
\caption{Performance comparison on gMission $100\times100$ for different $\rho$. \ouralg with $\rho=y$ means \ouralg is trained with $\rho=y$.}
\label{table:gmission_100by100}
\end{table}

We show the results for \ouralg and \rob under different $\rho\in[0,1]$ in
Table~\ref{table:gmission_100by100}.
In general, by setting a larger $\rho$ for inference, both \ouralg and \rob are closer to the expert algorithm \greedy. As \greedy works empirically much better than \ml in terms of the average performance and the worst-case performance, both \ouralg and \rob have better performances when we increase $\rho$ to let \greedy safeguard the RL decisions more aggressively. 
Moreover, by training the RL model with explicit awareness of online switching, \ouralg can have a higher average cost than \rob, which further demonstrates the benefits of training the RL model by considering its downstream operation. Also, interestingly,
by setting $\rho$ identical
for both training and testing, the average reward may not be the highest
for \ouralg, partially  
%This is partially 
 because
of the empirical testing dataset.
Another reason is that, in this test, \ml alone already performs very badly (both on average and in the worst case) due to the significant training-testing distributional discrepancy. Hence, by setting a higher $\rho$,
\ouralg  works better empirically though it is tested under a different $\rho$. An exception is  when testing 
\ouralg with $\rho=0.9$: setting $\rho=0.6/0.8$ for training makes \ouralg perform slightly better in
terms of the average performance and worst-case performance, respectively. 
But, setting $\rho=0.9$ for training still brings benefits to \ouralg compared to \rob that does not consider the downstream online switching operation.

\emph{To sum up}, our experimental results under different settings demonstrate: 
\ouralg's empirical improvement in terms of the average reward  compared to \rob;
the improved competitive ratio of \ouralg and \rob compared to \ml, especially when the training-testing distributions differ significantly;
and the improved average reward
of \ouralg compared to \greedy when
RL is good. 
Therefore, \ouralg can  exploit the power of RL while provably guaranteeing the performance robustness.

\section{Free Disposal}\label{appendix_free_disposal}

The {offline} version of  bipartite matching with free disposal can be expressed as:
\begin{equation}\label{eqn:offline_problem_free}
	\begin{split}
\textbf{With Free Disposal:}\;\;\;\;\;\;    &\max\sum_{x_{uv}\in\{0,1\}, u\in\mathcal{U}} \left(\max_{\mathcal{S}\subseteq\mathcal{V}_u, |\mathcal{S}|\leq c_u}\sum_{v\in\mathcal{S}}w_{uv}\right)\\
	\mathrm{s.t.}\;\;\;\;\;\;  & \mathcal{V}_u=\{v\in{\mathcal{V}\,|\,
	x_{uv}=1}\}\;\; \forall u\in\mathcal{U},\;\;\;	 \sum_{u\in\mathcal{U}}x_{uv}\leq 1,\;\; \forall v\in\mathcal{V},
\end{split}
\end{equation}
where $\mathcal{V}_u=\{v\in{\mathcal{V}\,|\,
	x_{uv}=1}\}$ is the set
	of online items matched to $u\in\mathcal{U}$
	and the objective $\max_{\mathcal{S}\in\mathcal{V}_u, |\mathcal{S}|\leq c_u}\sum_{v\in\mathcal{S}}x_{uv}w_{uv}$ indicates
	that only up to top $c_u$ rewards
are counted for $u\in\mathcal{U}$.

\begin{algorithm}[!t]
	\caption{Inference of
	  \ouralg(Free Disposal)} 
	\begin{algorithmic}[1]\label{alg:free_disposal}
		\STATE \textbf{Initialization:} 
		The actual set of  items
		matched to $u\in\mathcal{U}$ is $\mathcal{V}_{u,v}$ after sequentially-arriving item $v$'s assignment with
		$\mathcal{V}_{u,0}=\emptyset$, the actual remaining capacity is $b_u=c_u$ for  $u\in\mathcal{U}$,
		and the actual cumulative reward is $R_0=\sum_{u\in\mathcal{U}}f_{u}(\mathcal{V}_{u,0})=0$.
		The same notations apply to the expert algorithm $\pi$ by adding the superscript $\pi$. Competitive ratio requirement $\rho\in[0,1]$ and slackness $B\geq0$ with respect to the expert algorithm $\pi$.
		\FOR {$v= 1$ to $|\mathcal{V}|$}
		\STATE   Run the  algorithm $\pi$ and match
		the item $v$ to $u\in\mathcal{U}$
		based on the expert's decision $u=x_{v}^{\pi}$.
		\STATE Update the expert's decision set and reward for offline item $u=x_{v}^{\pi}$:\\
		$\mathcal{V}_{x_{v}^{\pi},v}^{\pi}=
		\mathcal{V}_{x_{v}^{\pi},v-1}^{\pi} \bigcup\{v\}$ and 
		$f_{x_{v}^{\pi}}=f_{x_{v}^{\pi}}(\mathcal{V}_{x_{v}^{\pi},v}^{\pi} )$.
		\STATE   Update the expert's cumulative reward 
		$R^{\pi}_v=\sum_{u\in\mathcal{U}}f_u$\\ 
        \FOR {$u$ in $\mathcal{U}$}
		\STATE Collect the available history information $I_{u}$ about item $u$ 
		\STATE Run the RL model to get score: ${s}_u=w_{uv}-h_{\theta}(I_{u},	w_{uv})$ where $\theta$ is the network weight
		\ENDFOR
		\STATE Calculate the probability of choosing each item $u$: $\left\lbrace \{\tilde{s}_u\}_{u\in\mathcal{U}}\right\rbrace =\mathrm{softmax}\left\lbrace \{{s}_u\}_{u\in\mathcal{U}}\right\rbrace $.
		\STATE Obtain RL decision: $\tilde{x}_v=\arg\max_{u\in\mathcal{U}\bigcup \{\mathrm{skip}\}}\left\lbrace \{\tilde{s}_u\}_{u\in\mathcal{U}}\right\rbrace$.
		\STATE Find 
		$\Delta f_{\tilde{x}_v}$ in Eqn.~\eqref{eqn:additional_reward_free}
		and $G\left(\tilde{x}_v,\{\mathcal{V}_{u,v-1}\}_{u\in\mathcal{U}},\{\mathcal{V}_{u,v}^{\pi}\}{u\in\mathcal{U}}\right)$ in Eqn.~\eqref{eqn:hedging_free}
		\IF{$R_{v-1}+
		\Delta f_{\tilde{x}_v}
		\geq \rho \left( R^{\pi}_v +G\left(\tilde{x}_v,\{\mathcal{V}_{u,v-1}\}_{u\in\mathcal{U}},\{\mathcal{V}_{u,v}^{\pi}\}_{u\in\mathcal{U}}\right)\right)-B $}
		\STATE Select $x_v=\tilde{x}_v$.  \quad  \texttt{//Follow the ML action}
		\ELSE 
		\STATE Select $x_v=x^{\pi}_{v}$.  \quad  \texttt{//Follow the expert}
		\ENDIF
		\STATE Update assignment and
		 reward: 
		 $\mathcal{V}_{x_{v},v}=
		\mathcal{V}_{x_{v},v-1}\bigcup\{v\}$ and
		 $R_v=\sum_{u\in\mathcal{U}}f_{u}(\mathcal{V}_{u,v})$ 
		\ENDFOR
	\end{algorithmic}
\end{algorithm}

In the free-disposal setting, it is more challenging to design the switching conditions to guarantee the robustness. The reason is the additional flexibility
allowed for matching decisions --- each offline item
$u\in\mathcal{U}$ is allowed to be matched more than $c_u$ times although only
up to top $c_u$ rewards actually count \cite{OnlineMatching_Booklet_OnlineMatchingAdAllocation_Mehta_2013_41870,OBM_EdgeWeighted_OBM_ZhiyiHuang_FOCS_2020_9317873}. For example, even though \ouralg and the expert assign the same number of online items to 
an offline item $u\in\mathcal{U}$
and \ouralg is better than the expert at a certain step, future  high-reward online items can still be assigned to $u\in\mathcal{U}$, increasing the expert's total reward 
or even equalizing the rewards
of \ouralg and the expert (i.e.,
high-reward future online items become the top $c_u$ items for $u\in\mathcal{U}$ for both \ouralg and the expert). Thus, the temporarily ``higher'' rewards
received by \ouralg must be hedged against such future uncertainties.
Before designing our switching condition for the free-disposal setting, we first
define the set 
containing the top $c_u$ online items for $u\in\mathcal{U}$ after assignment
of $v$:
\begin{equation}\label{eqn:define_e}
    \mathcal{E}_{u,v}(\mathcal{V}_{u,v})=\arg\max_{\mathcal{E}\subseteq\mathcal{V}_{u,v}, |\mathcal{E}|= c_u}\sum_{v\in\mathcal{E}}w_{uv},
\end{equation}
where $\mathcal{V}_{u,v}$ is the set of all online items matched to $u\in\mathcal{U}$ so far after assignment of item $v\in\mathcal{V}$. When there are fewer than $c_u$ items in $\mathcal{V}_{uv}$, we will simply add null items with reward $0$ to $\mathcal{E}_{u,v}$ such that $|\mathcal{E}_{u,v}|=c_{u}$.
 %to make it have $c_u$ items.
We also sort the online items denoted as $e_{u,i}$, for $i=1,\cdots, c_u$,
contained in $\mathcal{E}_{u,v}$
according to their weights in an increasing order such
that $w_{u,e_{u,1}}\leq \cdots\leq w_{u,e_{u,c_u}}$. Similarly,
we define the same top-$c_u$ item set for the expert algorithm $\pi$ by adding
the superscript $\pi$.

Next, we define the following value which indicates the maximum possible additional reward  
for the expert algorithm $\pi$ 
if \ouralg simply switches to the expert and follows it for all the future steps $v+1,v+2,\cdots$:
\begin{equation}\label{eqn:hedging_free_proof}
    G\left(\tilde{x}_v,\{\mathcal{V}_{u,v-1}\}_{u\in\mathcal{U}},\{\mathcal{V}_{u,v}^{\pi}\}_{u\in\mathcal{U}}\right)=\sum_{u\in\mathcal{U}}
    \left(\max_{i=1,\cdots,c_u}\sum_{j=1}^i (w_{u,e_{u,j}}-w_{u,e_{u,j}^{\pi}})\right)^+,
\end{equation}
where $e_{u,j}^{\pi}\in\mathcal{E}_{u}^{\pi}({\mathcal{V}}_{u,v}^{\pi})$, and 
$e_{u,j}\in\mathcal{E}_{u}(\tilde{\mathcal{V}}_{u,v})$ in which  
$\tilde{\mathcal{V}}_{u,v}=\mathcal{V}_{u,v-1}$ if $\tilde{x}_v\not=u$
and $\tilde{\mathcal{V}}_{u,v}=\mathcal{V}_{u,v-1}\bigcup\{v\}$ if $\tilde{x}_v=u$.

The interpretation is as follows. Suppose that 
\ouralg follows the RL decision for online item $v$.
If it has a higher cumulative
reward for the $j$-th item in the top-$c_u$ item set $\mathcal{E}_{u,v}$
than the expert algorithm $\pi$, then the expert can still possibly
offset the reward difference $w_{u,e_{u,j}}-w_{u,e_{u,j}^{\pi}}$
by receiving a high-reward future online item that replaces the $j$-th item
for both \ouralg and the expert.
Nonetheless, in the free-disposal model,
the items in the top-$c_u$ set $\mathcal{E}_{u,v}$ are removed sequentially --- the lowest-reward
item will be first removed from the sorted set $\mathcal{E}_{u,v}$, followed by
the next lowest-reward item, and so on. 
Thus, in order for a high-reward item to replace the $i$-th item
in the sorted set $\mathcal{E}_{u,v}$,
the first $(i-1)$ items have to be removed first by other high-reward online items. As a result, if
\ouralg has a lower
reward for the $j$-th item (for $j\leq i$) in the top-$c_u$ item set $\mathcal{E}_{u,v}$
than the expert algorithm $\pi$, then it will negatively impact the expert's additional reward gain in the future. Therefore, for item $u\in\mathcal{U}$
we only need to find the highest total reward difference,
$\left(\max_{i=1,\cdots,c_u}\sum_{j=1}^i (w_{u,e_{u,j}}-w_{u,e_{u,j}^{\pi}})\right)^+$,
that can be offset
for the expert algorithm $\pi$ by
considering that $i$ items are replaced by future high-reward online items for $i=1,\cdots, c_u$.
If $\max_{i=1,\cdots,c_u}\sum_{j=1}^i (w_{u,e_{u,j}}-w_{u,e_{u,j}^{\pi}})$ is negative (i.e., the expert algorithm cannot possibly gain higher rewards than \ouralg by receiving high-reward online items to replace its existing ones), then we use $0$ as the hedging reward.

Finally, by summing up the hedging
rewards for all the offline items $u\in\mathcal{U}$, we obtain the total hedging reward in Eqn.~\eqref{eqn:hedging_free_proof}.
Based on this hedging reward, we have the condition (Line~28 in Algorithm~\ref{alg:1} for \ouralg to follow the RL decision:
$R_{v-1}+
		\Delta f_{\tilde{x}_v}
		\geq \rho \left( R^{\pi}_v +G\left(\tilde{x}_v,\{\mathcal{V}_{u,v-1}\}_{u\in\mathcal{U}},\{\mathcal{V}_{u,v}^{\pi}\}_{u\in\mathcal{U}}\right)\right)-B $,
where $\Delta f_{\tilde{x}_v}$ defined below is the additional reward if $\tilde{x}_v$ is not skip, which would be obtained by following the RL decision:
\begin{equation}\label{eqn:additional_reward_free}
  \Delta f_{\tilde{x}_v}=
  f_{\tilde{x}_v}(\mathcal{V}_{\tilde{x}_v,v}\bigcup\{v\})-	f_{\tilde{x}_v}(\mathcal{V}_{\tilde{x}_v,v-1}),  
\end{equation}
in which 
$f_u=f_u(\mathcal{V}')= \max_{\mathcal{S}\in\mathcal{V}', |\mathcal{S}|\leq c_u}\sum_{v\in\mathcal{S}}w_{uv}$ is the reward function for an offline item $u\in\mathcal{U}$ in the free-disposal model. The condition means that if \ouralg can maintain
the competitive ratio $\rho$ against the expert algorithm $\pi$ by   
being able to hedge against any future uncertainties even in the worst case,
then it can safely follow the RL decision $\tilde{x}_v$ at step $v$.

\textbf{Training with free disposal.} The training process for the free-disposal setting is the same
as that for the no-free-disposal setting, except for we need to modify reward difference $R_{diff}$
based on the switching condition 
(i.e., Line~13 of Algorithm~\ref{alg:free_disposal}) for the free-disposal setting. The $R_{diff}$ is obtained by subtracting right hand side from the left hand side of the switching condition, which is used to calculate $p_{\theta}(x_v \mid I_u)$ in Line~3 of Algorithm~\ref{alg:training_brief}.

\section{Proof of Theorem~\ref{thm:cr}}\label{appendix_proof}

The key idea of proving Theorem~\ref{thm:cr} is to show
that there always exist feasible actions (either following the expert
or skip) while being able to guarantee the robustness if we follow the switching condition. Next, we prove Theorem~\ref{thm:cr} for the no-free-disposal and free-disposal settings, respectively.

\subsection{No Free Disposal}
Denote $\mathcal{V}_{u,v}$ as the actual set of items matched to $u\in\mathcal{U}$ after making decision for $v$. Denote $\mathcal{V}_{u,v}^\pi$ as the expert's set of items matched to $u\in\mathcal{U}$. We first prove a technical lemma.

\begin{lemma}\label{lma:hold}
Assuming that the robustness condition is met after making the decision for $v-1$, i.e. $R_{v-1}\geq \rho \left( R^{\pi}_{v-1} + \sum_{u\in \mathcal{U}}\left(|\mathcal{V}_{u,v-1}|-|V_{u,v-1}^{\pi}| \right)^+\cdot w_{u,\max} \right)-B$.
If at the step when $v$ arrives and the expert's decision $x^{\pi}_v$ is not available for matching, then $x_v=\mathrm{skip}$ always satisfies $R_{v}\geq \rho \left( R^{\pi}_{v} + \sum_{u\in \mathcal{U}}\left(|\mathcal{V}_{u,v}|-|V_{u,v}^{\pi}| \right)^+\cdot w_{u,\max} \right)-B$.
\end{lemma}
\begin{proof} 

If the item $x^{\pi}_v$ is not available for matching, it must have been consumed before $v$ arrives, which means 
$|\mathcal{V}_{x^{\pi}_v,v-1}|-|V_{x^{\pi}_v,v-1}^{\pi}| \geq 1$ (since otherwise
the expert cannot choose $x_v^{pi}$ either).
Since $x_v=\mathrm{skip}$, we have $R_v=R_{v-1}$ and $\mathcal{V}_{u,v} = \mathcal{V}_{u,v-1}, \quad \forall u\in \mathcal{U}$. 
Then, by the robustness assumption of the previous step, we have 
\begin{equation}
\begin{split}
R_v=R_{v-1} \geq & \rho \left( R^{\pi}_{v-1} + \sum_{u\in \mathcal{U}}\left(|\mathcal{V}_{u,v-1}|-|V_{u,v-1}^{\pi}| \right)^+\cdot w_{u,\max} \right)-B\\
 \geq & \rho \left( R^{\pi}_{v-1}+w_{x^{\pi}_v,v} -w_{x^{\pi}_v,\max} + \sum_{u\in \mathcal{U}}\left(|\mathcal{V}_{u,v-1}|-|V_{u,v-1}^{\pi}| \right)^+\cdot w_{u,\max} \right)-B\\
 = &\rho\left( R^{\pi}_{v} + \sum_{u\in \mathcal{U}}\left(|\mathcal{V}_{u,v}|-|V_{u,v}^{\pi}| \right)^+\cdot w_{u,\max}  \right)-B
\end{split}
\end{equation}
where the last equality holds because $R^{\pi}_{v} = R^{\pi}_{v-1}+w_{x^{\pi}_v,v}$, and $(|\mathcal{V}_{u,v}|-|\mathcal{V}_{u,v}^{\pi}|)^+ - (|\mathcal{V}_{u,v-1}|-|\mathcal{V}_{u,v-1}^{\pi}|)^+ = -1$ if $u = x_v^\pi$, and $(|\mathcal{V}_{u,v}|-|\mathcal{V}_{u,v}^{\pi}|)^+ - (|\mathcal{V}_{u,v-1}|-|\mathcal{V}_{u,v-1}^{\pi}|)^+ = 0$ otherwise.
\end{proof}

Next we prove by induction that the condition 
\begin{equation}\label{eqn:conditionafter}
R_{v}\geq \rho \left( R^{\pi}_{v} + \sum_{u\in \mathcal{U}}\left(|\mathcal{V}_{u,v}|-|V_{u,v}^{\pi}| \right)^+\cdot w_{u,\max} \right)-B
\end{equation}
holds for all steps by Algorithm \ref{alg:1}.

At the first step, if $\tilde{x}_v$ is not the same as $x^{\pi}_v$ and $w_{\tilde{x}_v,v}\geq \rho \left( w_{x^{\pi}_v,v}+w_{u,\max}\right)-B $, we select the RL decision $x_v=\tilde{x}_v$, and the robustness condition \eqref{eqn:conditionafter} is satisfied. Otherwise, we select the expert action $x_v=x_v^{\pi}$ and the condition still holds since the reward is non-negative, 
%$A_v=C_v$ and $w_{x_v,v}\geq \rho  w_{x^{\pi}_v,v}-B $ holds when 
$\rho\leq 1$ and $B\geq 0$. 

Then, assuming that the robustness condition in \eqref{eqn:conditionafter} is satisfied after making the decision for $v-1$, we need to prove it is also satisfied after making the decision for $v$. If the condition in \eqref{eqn:condition} in Algorithm~\ref{alg:1} is satisfied, then $x_v=\tilde{x}_v$ and so \eqref{eqn:conditionafter} holds naturally. 
Otherwise, if the expert action $x^{\pi}_v$ is available for matching, then we select expert action $x_v=x_v^{\pi}$. Then, we have $w_{x^{\pi}_v,v} \geq 0$ and $|\mathcal{V}_{u,v}|-|V_{u,v}^{\pi}| = |\mathcal{V}_{u,v-1}|-|V_{u,v-1}^{\pi}|, \quad \forall u\in \mathcal{U}$,  hence the condition \eqref{eqn:conditionafter} still holds. Other than these two cases, we also have the option to ``skip'', i.e. $x_v=\mathrm{skip}$. By Lemma~\ref{lma:hold}, the condition \eqref{eqn:conditionafter} still holds. Therefore, we prove that the condition~\eqref{eqn:conditionafter} holds for every step. 

After the last step $v=|\mathcal{V}|$, we must have 
\begin{equation}
\begin{split}
R_{{v}} \geq \rho \left( R^{\pi}_{{v}} + \sum_{u\in \mathcal{U}}\left(|\mathcal{V}_{u,\hat{v}}|-|V_{u,\hat{v}}^{\pi}| \right)^+\cdot w_{u,\max} \right)-B\geq \rho R_{{v}}^{\pi} -B 
\end{split}
\end{equation}
where $R_{{v}}$ and $R^{\pi}_{{v}}$
are the total rewards of \ouralg 
and the expert algorithm $\pi$ after the last step $v=|\mathcal{V}|$, respectively.
This completes the proof for the no-free-disposal case.

\subsection{With Free Disposal}
We now turn to the free-disposal setting which is more challenging than the no-free-disposal setting because of the possibility of using future high-reward items to replace existing low-reward ones. 

We first denote $\Delta f_{{x}_v^\pi}$ as the actual additional reward obtained by following the expert's decision ${x}_v^\pi$, 
\begin{equation}
  \Delta f_{x_v^\pi}=
  f_{x_v^\pi}(\mathcal{V}_{x_v^\pi,v}\bigcup\{v\})-	f_{x_v^\pi}(\mathcal{V}_{x_v^\pi,v-1}),  
\end{equation}
Additionally, we denote $\Delta f_{x_v^\pi}^\pi$ as the expert's additional reward of choosing ${x}_v^\pi$, where
\begin{equation}
  \Delta f_{x_v^\pi}^\pi=
  f_{x_v^\pi}(\mathcal{V}_{x_v^\pi,v}^\pi\bigcup\{v\})-	f_{x_v^\pi}(\mathcal{V}_{x_v^\pi,v-1}^\pi).
\end{equation}

For presentation convenience, we rewrite the hedging reward as $\tilde{G}\left(\{\mathcal{V}_{u,v}\}_{u\in\mathcal{U}},\{\mathcal{V}_{u,v}^{\pi}\}_{u\in\mathcal{U}}\right)$ as 
\begin{equation}\label{eqn:hedging_free}
    \tilde{G}\left(\{\mathcal{V}_{u,v}\}_{u\in\mathcal{U}},\{\mathcal{V}_{u,v}^{\pi}\}_{u\in\mathcal{U}}\right)=\sum_{u\in\mathcal{U}}
    \left(\max_{i=1,\cdots,c_u}\sum_{j=1}^i (w_{u,e_{u,j}}-w_{u,e_{u,j}^{\pi}})\right)^+,
\end{equation}
where $e_{u,j}^{\pi}\in\mathcal{E}_{u}^{\pi}({\mathcal{V}}_{u,v}^{\pi})$, 
$e_{u,j}\in\mathcal{E}_{u}({\mathcal{V}}_{u,v})$, and $\mathcal{E}_{u}$ is defined in Eqn.~\eqref{eqn:define_e}.

\begin{lemma}\label{lemma:with_disposal}
Assuming that the robustness condition is met after making the decision for $v-1$, i.e. 
$R_{v-1} \geq \rho \left( R^{\pi}_{v-1} +\tilde{G}\left(\{\mathcal{V}_{u,v-1}\}_{u\in\mathcal{U}},\{\mathcal{V}_{u,v-1}^{\pi}\}_{u\in\mathcal{U}}\right) \right)-B$.
At step $v$, we have $\Delta f_{{x}_v^\pi} - \Delta f_{x_v^\pi}^\pi \geq G\left(x_v^\pi,\{\mathcal{V}_{u,v-1}\}_{u\in\mathcal{U}},\{\mathcal{V}_{u,v}^{\pi}\}_{u\in\mathcal{U}}\right) - \tilde{G}\left(\{\mathcal{V}_{u,v-1}\}_{u\in\mathcal{U}},\{\mathcal{V}_{u,v-1}^{\pi}\}_{u\in\mathcal{U}}\right)$.
\end{lemma}

\begin{proof}
We begin with 
``$G\left(x_v^\pi,\{\mathcal{V}_{u,v-1}\}_{u\in\mathcal{U}},\{\mathcal{V}_{u,v}^{\pi}\}_{u\in\mathcal{U}}\right) - \tilde{G}\left(\{\mathcal{V}_{u,v-1}\}_{u\in\mathcal{U}},\{\mathcal{V}_{u,v-1}^{\pi}\}_{u\in\mathcal{U}}\right)$''
in
Lemma~\ref{lemma:with_disposal}.
By definition, it can be written as
\begin{equation} 
\begin{split}\label{eqn:reserve_free_dp}
 & G\left(x_v^\pi,\{\mathcal{V}_{u,v-1}\}_{u\in\mathcal{U}},\{\mathcal{V}_{u,v}^{\pi}\}_{u\in\mathcal{U}}\right) - \tilde{G}\left(\{\mathcal{V}_{u,v-1}\}_{u\in\mathcal{U}},\{\mathcal{V}_{u,v-1}^{\pi}\}_{u\in\mathcal{U}}\right) \\
 = & \left(\max_{i=1,\cdots,c_u}\sum_{j=1}^i (w_{u,\hat{e}_{u,j}}-w_{u,\hat{e}_{u,j}^{\pi}})\right)^+ - \left(\max_{i=1,\cdots,c_u}\sum_{j=1}^i (w_{u,{e}_{u,j}}-w_{u,{e}_{u,j}^{\pi}})\right)^+
\end{split}
\end{equation}

where $u = x_v^\pi$, $\hat{e}_{u,j}^{\pi}\in\mathcal{E}_{u}^{\pi}(\mathcal{V}_{u,v-1}^\pi\bigcup\{v\} )$, and 
$\hat{e}_{u,j}\in \mathcal{E}_{u}(\mathcal{V}_{u,v-1}\bigcup\{v\})$. 
Besides, ${e}_{u,j}^{\pi}\in\mathcal{E}_{u}^{\pi}({\mathcal{V}}_{u,v-1}^{\pi})$, and ${e}_{u,j}\in\mathcal{E}_{u}(\mathcal{V}_{u,v-1})$.

To prove the lemma, we consider four possible cases for $w_{u,v}$ to
cover all the cases.

\textbf{Case 1}: If the reward for $v$ is small enough such that $w_{u,v} < w_{u, e_{u,1}}$ and $w_{u,v} < w_{u, e_{u,1}^\pi}$, then $v \notin \mathcal{E}_{u}(\mathcal{V}_{u,v-1}\bigcup\{v\})$ and $v \notin \mathcal{E}_{u}(\mathcal{V}^\pi_{u,v-1}\bigcup\{v\})$. Then we have $\Delta f_{{x}_v^\pi} = \Delta f_{x_v^\pi}^\pi = 0$, since both the expert and \ouralg cannot gain any  reward from the online item $v$. From Eqn.~\eqref{eqn:reserve_free_dp}, we can find that the right-hand side is also 0. Therefore, the conclusion in Lemma~\ref{lemma:with_disposal} 
 holds with the equality activated. 

\textbf{Case 2}: If the reward for $v$ is large enough such that $w_{u,v} > w_{u, e_{u,1}}$ and $w_{u,v} > w_{u, e_{u,1}^\pi}$, then $v \in \mathcal{E}_{u}(\mathcal{V}_{u,v-1}\bigcup\{v\})$ and $v \in \mathcal{E}_{u}(\mathcal{V}^\pi_{u,v-1}\bigcup\{v\})$. In other words, we will remove the smallest-reward item $ e_{u,1} \notin \mathcal{E}_{u}(\mathcal{V}_{u,v-1}\bigcup\{v\})$ and $e_{u,1}^\pi \notin \mathcal{E}_{u}(\mathcal{V}^\pi_{u,v-1}\bigcup\{v\})$. Then
$$G\left(x_v^\pi,\{\mathcal{V}_{u,v-1}\}_{u\in\mathcal{U}},\{\mathcal{V}_{u,v}^{\pi}\}_{u\in\mathcal{U}}\right) - \tilde{G}\left(\{\mathcal{V}_{u,v-1}\}_{u\in\mathcal{U}},\{\mathcal{V}_{u,v-1}^{\pi}\}_{u\in\mathcal{U}}\right)  \leq  -w_{u, e_{u,1}} + w_{u, e_{u,1}^\pi} $$
The inequality holds because $(w_{u, e_{u,1}} - w_{u, e_{u,1}^\pi})^+ \geq w_{u, e_{u,1}} - w_{u, e_{u,1}^\pi}$. In this case, $\Delta f_{{x}_v^\pi} = w_{u,v} - w_{u, e_{u,1}}$ and $\Delta f_{{x}_v^\pi}^\pi = w_{u,v} - w_{u, e_{u,1}^\pi}$. Therefore, the conclusion in Lemma~\ref{lemma:with_disposal} 
 holds. 

\textbf{Case 3}: If the reward for $v$ satisfies $w_{u,v} \geq w_{u, e_{u,1}}$ and $w_{u,v} \leq w_{u, e_{u,1}^\pi}$, then $v \in \mathcal{E}_{u}(\mathcal{V}_{u,v-1}\bigcup\{v\})$ and $v \notin \mathcal{E}_{u}(\mathcal{V}^\pi_{u,v-1}\bigcup\{v\})$.
In other words, 
even if $v \in \mathcal{E}_{u}(\mathcal{V}_{u,v-1}\bigcup\{v\})$
(i.e., the online item $v$ produces additional rewards for \ouralg), 
the reward of $v$ is still smaller than the smallest reward for the expert.
Then, only the lowest reward of \ouralg will be kicked out. Then we have $G\left(x_v^\pi,\{\mathcal{V}_{u,v-1}\}_{u\in\mathcal{U}},\{\mathcal{V}_{u,v}^{\pi}\}_{u\in\mathcal{U}}\right) - \tilde{G}\left(\{\mathcal{V}_{u,v-1}\}_{u\in\mathcal{U}},\{\mathcal{V}_{u,v-1}^{\pi}\}_{u\in\mathcal{U}}\right) \leq w_{u,v} - w_{u, e_{u,1}} $, the equality activates if $G\left(x_v^\pi,\{\mathcal{V}_{u,v-1}\}_{u\in\mathcal{U}},\{\mathcal{V}_{u,v}^{\pi}\}_{u\in\mathcal{U}}\right) \geq 0$. 
In this case, $\Delta f_{{x}_v^\pi} = w_{u,v} - w_{u, e_{u,1}}$ and $\Delta f_{{x}_v^\pi}^\pi = 0$. Therefore,
the conclusion in Lemma~\ref{lemma:with_disposal} 
still holds. 

\textbf{Case 4}: If the reward for $v$ satisfies $w_{u,v} \leq w_{u, e_{u,1}}$ and $w_{u,v} \geq w_{u, e_{u,1}^\pi}$, then in this case, only the current smallest-reward item is replaced with $v$ for the expert, 
while the reward of \ouralg remains unchanged. Thus, we have
$$G\left(x_v^\pi,\{\mathcal{V}_{u,v-1}\}_{u\in\mathcal{U}},\{\mathcal{V}_{u,v}^{\pi}\}_{u\in\mathcal{U}}\right) - \tilde{G}\left(\{\mathcal{V}_{u,v-1}\}_{u\in\mathcal{U}},\{\mathcal{V}_{u,v-1}^{\pi}\}_{u\in\mathcal{U}}\right)  =   w_{u, e_{u,1}^\pi} - w_{u, v}.$$
In this case, $\Delta f_{{x}_v^\pi} = 0$ and $\Delta f_{{x}_v^\pi}^\pi = w_{u, v} - w_{u, e_{u,1}^\pi}$. Then the conclusion in Lemma~\ref{lemma:with_disposal} 
still holds with the equality activated.
\end{proof}

We next prove by induction that the condition 
\begin{equation}\label{eqn:condition_free_dis}
R_{v} \geq \rho \left( R^{\pi}_v + \tilde{G}\left(\{\mathcal{V}_{u,v}\}_{u\in\mathcal{U}},\{\mathcal{V}_{u,v}^{\pi}\}_{u\in\mathcal{U}}\right) \right)-B
\end{equation}
holds for all steps by Algorithm \ref{alg:1}.

At the first step, by using ${x}_v = x_v^\pi$, we have $R_v = R_v^\pi$ and $\tilde{G}\left(\{\mathcal{V}_{u,v}\}_{u\in\mathcal{U}},\{\mathcal{V}_{u,v}^{\pi}\}_{u\in\mathcal{U}}\right) = 0$,
and it is obvious that the condition in \eqref{eqn:condition_free_dis} is satisfied. Thus, there is at least one solution ${x}_v = x_v^\pi$ for our robustness condition in  \eqref{eqn:condition_free_dis}. 

Starting from the second step, assume that after the step $v-1$, we already have 
\begin{equation}\label{eqn:disposal_step}
    R_{v-1} \geq \rho \left( R^{\pi}_{v-1} + \tilde{G}\left(\{\mathcal{V}_{u,v-1}\}_{u\in\mathcal{U}},\{\mathcal{V}_{u,v-1}^{\pi}\}_{u\in\mathcal{U}}\right) \right)-B
\end{equation}
If the condition in Line~13 of Algorithm~\ref{alg:free_disposal} is already satisfied, we can just use $x_v = \tilde{x}_v$, which  directly satisfies \eqref{eqn:condition_free_dis}. Otherwise, we need to follow the expert by setting $x_v = x_v^\pi$. 
Now we will prove $x_v = x_v^\pi$ satisfies the robustness condition at any step $v$.

From Lemma~\ref{lemma:with_disposal}, since $0 \leq \rho \leq 1$ and $\Delta f_{{x}_v^\pi} \geq 0$ we have
$$\Delta f_{{x}_v^\pi}  \geq \rho\Delta f_{{x}_v^\pi} \geq \rho \left( \Delta f_{x_v^\pi}^\pi +  G\left(x_v^\pi,\{\mathcal{V}_{u,v-1}\}_{u\in\mathcal{U}},\{\mathcal{V}_{u,v}^{\pi}\}_{u\in\mathcal{U}}\right) - \tilde{G}\left(\{\mathcal{V}_{u,v-1}\}_{u\in\mathcal{U}},\{\mathcal{V}_{u,v-1}^{\pi}\}_{u\in\mathcal{U}}\right) \right).$$
Then, by substituting it back to Eqn.~\eqref{eqn:disposal_step}, we have
\begin{equation} 
\begin{split}
 R_{v-1} + \Delta f_{{x}_v^\pi} \geq& \rho \left(\Delta f_{x_v^\pi}^\pi + G\left(x_v^\pi,\{\mathcal{V}_{u,v-1}\}_{u\in\mathcal{U}},\{\mathcal{V}_{u,v}^{\pi}\}_{u\in\mathcal{U}}\right) - \tilde{G}\left(\{\mathcal{V}_{u,v-1}\}_{u\in\mathcal{U}},\{\mathcal{V}_{u,v-1}^{\pi}\}_{u\in\mathcal{U}}\right) \right)\\
 & +\rho \left( R^{\pi}_{v-1} + \tilde{G}\left(\{\mathcal{V}_{u,v-1}\}_{u\in\mathcal{U}},\{\mathcal{V}_{u,v-1}^{\pi}\}_{u\in\mathcal{U}}\right) \right)-B\\
 =& \rho \left( R^{\pi}_{v-1} + \Delta f_{x_v^\pi}^\pi + G\left(x_v^\pi,\{\mathcal{V}_{u,v-1}\}_{u\in\mathcal{U}},\{\mathcal{V}_{u,v}^{\pi}\}_{u\in\mathcal{U}}\right) \right)-B\\
 =& \rho \left( R^{\pi}_v + \tilde{G}\left(\{\mathcal{V}_{u,v}\}_{u\in\mathcal{U}},\{\mathcal{V}_{u,v}^{\pi}\}_{u\in\mathcal{U}}\right) \right)-B.
\end{split}
\end{equation}
Therefore, after the last step ${v}$, \ouralg must satisfy
$$R_{{{v}}} \geq \rho \left( R^{\pi}_{{v}} + \tilde{G}\left(\{\mathcal{V}_{u,{{v}}}\}_{u\in\mathcal{U}},\{\mathcal{V}_{u,{{v}}}^{\pi}\}_{u\in\mathcal{U}}\right) \right)-B \geq \rho R^{\pi}_{{v}} - B,$$
where $R_{{v}}$ and
$R^{\pi}_{{v}}$
are the total rewards of \ouralg and
the expert algorithm $\pi$ after the last step $v=|\mathcal{V}|$, respectively.
Thus, we complete the proof for the free-disposal setting.

\end{document}